\documentclass[lettersize,journal]{IEEEtran}
\usepackage{amsmath,amsfonts}
\usepackage{array}
\usepackage[caption=false,font=normalsize,labelfont=sf,textfont=sf]{subfig}
\usepackage{textcomp}
\usepackage{stfloats}
\usepackage{url}
\usepackage{verbatim}
\usepackage{multicol}
\usepackage{multirow}
\usepackage{graphicx}
\usepackage{cite}
\usepackage{url}
\usepackage{graphicx}
\usepackage{multirow}
\usepackage{booktabs} 
\usepackage[marginal]{footmisc} 
\usepackage{tipa} 
\usepackage{float} 
\usepackage{placeins}
\usepackage{bm} 
\usepackage{lipsum}
\usepackage{array} 
\usepackage{enumerate}
\usepackage{hyperref}
\usepackage{subfloat}
\usepackage{amsmath,bm}

\usepackage{diagbox}
\usepackage{booktabs}
\usepackage{amssymb}
\usepackage{balance} 
\usepackage{bbm} 
\usepackage{amsmath,amssymb,amsfonts}
\usepackage{graphicx}
\usepackage{textcomp}
\usepackage{xcolor}
\usepackage{algorithm}
\usepackage{algpseudocode}

\newcommand{\xy}[1]{\textcolor{black}{#1}}
\newcommand{\xyl}[1]{\textcolor{black}{#1}}

\begin{document}

\title{Harnessing Lightweight Transformer with Contextual Synergic Enhancement for Efficient \\ 3D Medical Image Segmentation}

\author{Xinyu Liu, Zhen Chen, \textit{Member, IEEE}, Wuyang Li, Chenxin Li, Yixuan Yuan, \textit{Senior Member, IEEE}
\thanks{\quad Xinyu Liu, Wuyang Li, Chenxin Li, and Yixuan Yuan are with the Department of Electronic Engineering, The Chinese University of Hong Kong, Hong Kong SAR (email: xinyuliu@link.cuhk.edu.hk; wymanbest@outlook.com; chenxinli@link.cuhk.edu.hk; yxyuan@ee.cuhk.edu.hk).}
\thanks{\quad Zhen Chen is with the Centre for Artificial Intelligence and Robotics (CAIR),
Hong Kong Institute of Science \& Innovation,
Chinese Academy of Sciences, Hong Kong SAR (email: zhen.chen@cair-cas.org.hk).}
\thanks{\quad \textit{Corresponding author: Yixuan Yuan (email: yxyuan@ee.cuhk.edu.hk).} This work was supported by Hong Kong Research Grants Council (RGC) General Research Fund under grant 14220622 and Hong Kong Innovation and Technology Commission Innovation and Technology Fund PRP/082/24FX.}
}

\markboth{Journal of \LaTeX\ Class Files,~Vol.~14, No.~8, August~2021}%
{Shell \MakeLowercase{\textit{et al.}}: A Sample Article Using IEEEtran.cls for IEEE Journals}


\maketitle

\begin{abstract}
Transformers have shown remarkable performance in 3D medical image segmentation, but their high computational requirements and need for large amounts of labeled data limit their applicability. To address these challenges, we consider two crucial aspects: model efficiency and data efficiency. Specifically, we propose Light-UNETR, a lightweight transformer designed to achieve model efficiency. Light-UNETR features a Lightweight Dimension Reductive Attention (LIDR) module, which reduces spatial and channel dimensions while capturing both global and local features via multi-branch attention. Additionally, we introduce a Compact Gated Linear Unit (CGLU) to selectively control channel interaction with minimal parameters. Furthermore, we introduce a Contextual Synergic Enhancement (CSE) learning strategy, which aims to boost the data efficiency of Transformers. It first leverages the \textit{extrinsic contextual information} to support the learning of unlabeled data with Attention-Guided Replacement, then applies Spatial Masking Consistency that utilizes \textit{intrinsic contextual information} to enhance the spatial context reasoning for unlabeled data. Extensive experiments on various benchmarks demonstrate the superiority of our approach in both performance and efficiency. For example, with only 10\% labeled data on the \xy{Left Atrial Segmentation dataset}, our method surpasses BCP by 1.43\% Jaccard while drastically reducing the FLOPs by 90.8\% and parameters by 85.8\%. \xyl{Code is released at \url{https://github.com/CUHK-AIM-Group/Light-UNETR}.}
\end{abstract}


\begin{figure}
    \centering
    \includegraphics[width=0.495\textwidth]{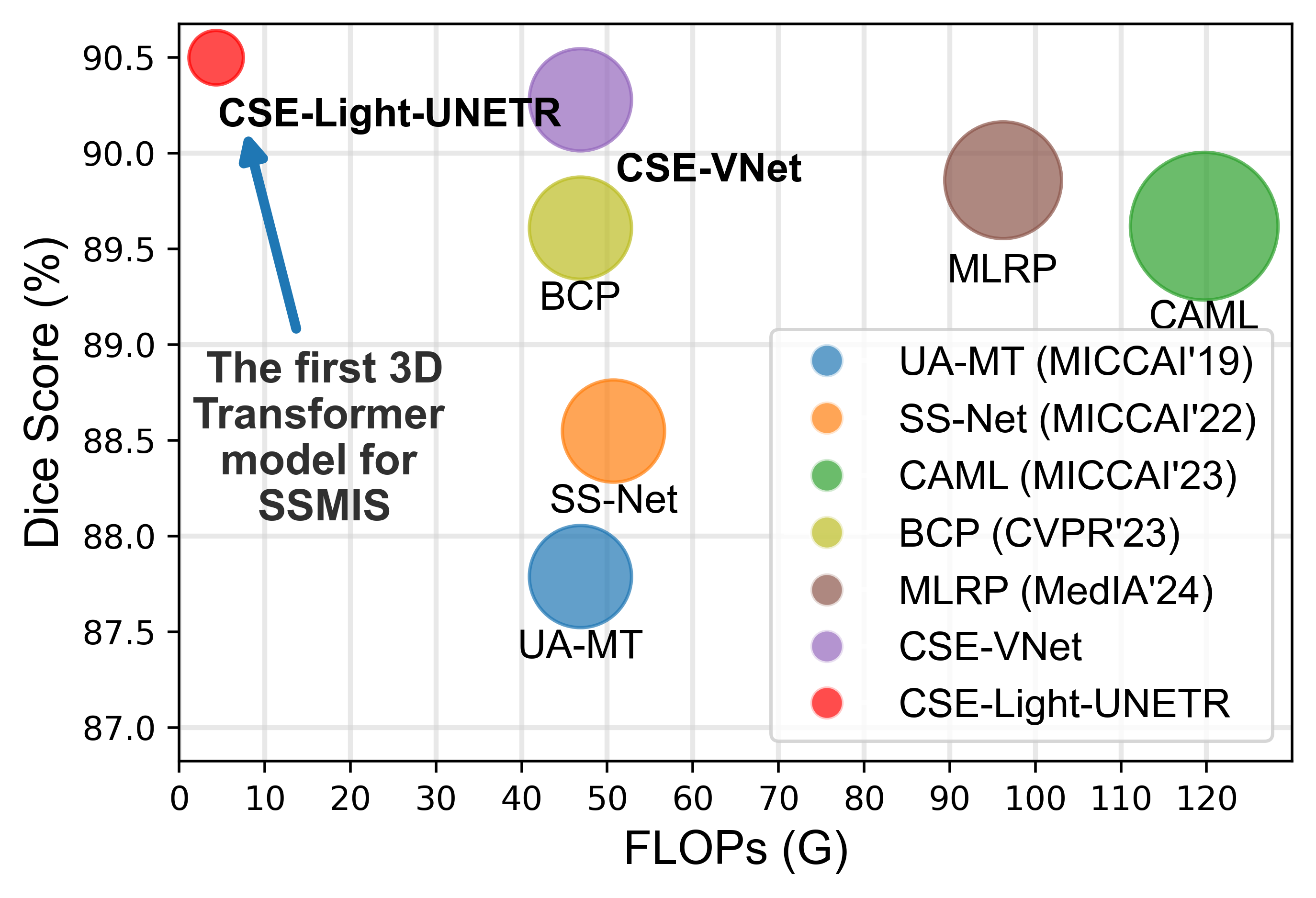}
    \caption{Performance comparison between the proposed CSE-Light-UNETR, CSE-VNet with previous SSMIS methods on the LA dataset \cite{LA}. The circle size represents the number of trainable parameters in different methods.}
    \label{fig:fig1_compare_all}
\end{figure}

\section{Introduction}
\label{sec:introduction}
Medical image segmentation is a crucial task in 3D medical imaging analysis, where the goal is to identify and separate different regions of interest in medical images such as Magnetic Resonance Imaging (MRI) or Computed Tomography (CT) \cite{chen2023adaptive, wang2018deepigeos, qiu2023rethinkingdualstream, li2025u}. This task is essential for various medical applications, including disease diagnosis, treatment planning, and image-guided interventions. 
However, the annotation process for 3D medical images is particularly challenging due to the inherent complexity of the data \cite{liu2023decoupled, yao2021label, LIU2021102052, wu2022minimizing, liu2022source}. Notably, annotating an object in 2D RGB image requires 40 seconds, whereas annotating a 3D CT volume demands a substantial 30-minute investment \cite{qu2024abdomenatlas}. This stark contrast underscores the pressing need for data efficient semi-supervised medical image segmentation (SSMIS) approaches \cite{yu2019uncertaintyawaremeanteacher, wu2022ssnet, xiang2022fussnet, BCP, CAML, wu2022minimizing, wu2022mutual, liu2024diffrect, liu2024most}, which can alleviate the annotation burden, 
and facilitate the deployment of deep learning models in medical imaging applications.

\xy{Traditional SSMIS approaches often rely on heavyweight and poorly performing CNN models such as V-Net \cite{vnet}, which can be computationally expensive and are incapable of learning global and more contextualized visual representations in medical scans. Despite their high computational cost, these models were favored for their robustness, reproducibility, and the ability to establish well-known benchmarks early in the field, providing a foundation for subsequent research. However, the emergence of lightweight architectures has challenged their dominance by offering more efficient solutions.}
In recent years, transformer-based models \cite{ViT, swin, cao2022swinunet, tang2022swinunetr, chen2021transunet, hatamizadeh2022unetr, wang2021transbts} have emerged as a promising alternative for medical image segmentation tasks. Owing to their ability to handle long-range dependencies, they could effectively tackle more complex and large-scale 3D medical image segmentation tasks, where traditional CNN-based approaches may struggle to scale. For example, UnetFormer \cite{hatamizadeh2022unetformer} and Swin UNETR \cite{tang2022swinunetr} pretrained their models on 5k CT images to achieve superior segmentation performance when finetuned on downstream datasets.
However, the application of transformers to SSMIS is hindered by several challenges. (1) Firstly, the massive parameter and computational requirements of medical segmentation transformers pose a significant obstacle to deployment in clinical practice and real-time diagnosis, as they typically necessitate multiple GPUs with substantial memory \cite{pang2023slimunetr, shaker2024unetr++}. This limitation is primarily attributed to the \textbf{lack of model efficiency}, which hinders the widespread adoption of transformers in medical imaging. (2) Furthermore, the transformer models often require enormous amounts of annotated data to achieve optimal performance, which can be a limiting factor in data-limited medical imaging applications, where annotated data is scarce. This \textbf{data efficiency bottleneck} needs innovative approaches that can mitigate the data requirements of transformers.

To enhance the model efficiency, we propose a lightweight transformer-based architecture, termed Light-UNETR. This architecture is built around the Light-UNETR block, which consists of two key components: Lightweight Dimension Reductive Attention (LIDR) and Compact Gated Linear Unit (CGLU). Considering that medical scans often exhibit complex anatomical structures and subtle lesions, they require the simultaneous capture of global contextual information and local fine-grained details. To this end, we design the LIDR block with multiple branches, incorporating self-attention mechanisms to capture global structures that correspond to low-frequency components, and multi-scale convolutions to extract local edges and textures that correspond to high-frequency components \cite{pan2022hilo, si2022inceptiontransformer, park2022vision}. Meanwhile, spatial and channel reduction are employed on the features to reduce the additional {computational costs}. 
Subsequently, the CGLU module selectively focuses on specific channels within the input feature, leveraging a gated mechanism to achieve {parameter efficiency}. By integrating these components, Light-UNETR is capable of capturing diverse and complex patterns in medical images while being significantly efficient.

Moreover, to enable transformers to learn medical scan segmentation in a data-efficient manner, we propose a semi-supervised learning framework termed Contextual Synergic Enhancement (CSE) that enables transformers to learn with limited annotated data. Unlike existing SSMIS methods that primarily rely on basic image and model regularization techniques to learn from unlabeled images \cite{wu2022mutual, luo2022semicnntransformer, sohn2020fixmatch},
CSE further captures complex extrinsic and intrinsic contextual relationships in both labeled and unlabeled medical volumes.
Specifically, CSE consists of two synergic components: an Attention-Guided Replacement strategy that selectively incorporates regions from labeled images into unlabeled images guided by attention values, thereby leveraging \textit{extrinsic contextual information} from labeled data to guide the understanding of unlabeled data; and a Spatial Masking Consistency strategy that encourages the model to reason the ambiguous regions via spatial context within the unlabeled data, which could utilize \textit{intrinsic contextual information} inherent in medical scans to infer the anatomy structure or lesions. 

Build upon the Light-UNETR and CSE, we establish the first pure 3D transformer-based SSMIS framework CSE-Light-UNETR, which achieves state-of-the-art performance and efficiency gains, as illustrated in Fig. \ref{fig:fig1_compare_all}. Notably, our Light-UNETR outperforms UNETR++ \cite{shaker2024unetr++} by 1.56\% in Dice score on the fully-supervised brain tumor segmentation task, while achieving a significant reduction in computational cost by 93.5\% of the total FLOPs. Moreover, in the semi-supervised scenario with 10\% labeled data, our CSE-Light-UNETR surpasses BCP \cite{BCP} by 1.43\% Jaccard while reducing the FLOPs by 90.8\% and parameters by 85.8\%. To summarize, our contributions are:

\begin{itemize}
    \item Considering the limited representation ability of CNN models and the large computational cost of transformers, we propose \textbf{a new lightweight transformer-based model Light-UNETR}, which is capable for achieving accurate and efficient segmentation of 3D medical images. It comprises LIDR that extract multi-granularity features, and CGLU that controls the channel interaction in a parameter-efficient manner.
    \item To address the need of massive data of transformers and fully leverage the contextual information in medical volumes, we propose \textbf{a data-efficient semi-supervised learning paradigm CSE}. It contains an Attention-Guided Replacement strategy that uses extrinsic context to infer the semantics in the unlabeled data, and a Spatial Masking Consistency strategy that leverages the intrinsic context in medical scans for spatial reasoning.
    \item We propose \textbf{an enhanced SSMIS framework by integrating CSE and Light-UNETR}, and conduct comprehensive semi- and fully-supervised experiments on diverse datasets encompassing multiple organs. Our approach achieves remarkable performance and efficiency trade-off, and surpasses existing methods significantly. 
\end{itemize}

\section{Related Work}
\label{sec:related}

\subsection{Transformers in Medical Image Segmentation}
The transformer \cite{vaswani2017attention} was originally proposed for natural language processing tasks, and has also demonstrated strong capability in capturing long-range dependencies in computer vision tasks \cite{ViT, deit, swin, liu2023efficientvit, si2022inceptiontransformer}. In recent years, transformers have been increasingly applied to medical image segmentation tasks. One of the earliest applications of transformers in medical image segmentation is TransUNet \cite{chen2021transunet}, which connects the Vision Transformer (ViT) architecture to a convolutional neural network (CNN) architecture, enhancing the information extraction ability of the encoder by extracting long-distance dependencies. 
Transfuse \cite{zhang2021transfuse} was proposed to use separate CNN and transformer branches to extract features and fuse them. This structure alleviates the depth of the network and the problem of ignoring intermediate features caused by the serial connection. 
Cao \textit{et al.} proposed SwinUNet \cite{cao2022swinunet}, which uses Swin Transformer \cite{swin} to extract middle features and designs a patch extension layer to complete up-sampling. This is the first medical image segmentation model based entirely on the transformer structure.

Although these models have made significant contributions to the field of medical image segmentation by leveraging the ability to extract long-distance dependencies of transformers, they are not directly applicable for 3D MRI or CT modalities. To address this drawback, UNETR \cite{hatamizadeh2022unetr}, UnetFormer \cite{hatamizadeh2022unetformer} and Swin UNETR \cite{hatamizadeh2021swinunetr} served as pioneering works for volumetric medical image segmentation with Transformers. Slim UNETR \cite{pang2023slimunetr} enabled information exchange via self-attention mechanism decomposition and an effective representation aggregation. UNETR++ \cite{shaker2024unetr++} further proposed an efficient paired attention block to learn spatial and channel-wise discriminative features. However, it is challenging to train a well-performed 3D transformer model, as they are data-hungry and require a large amount of data for pretraining \cite{hatamizadeh2022unetr}. Otherwise, they could suffer from training instability and collapse. Moreover, the transformers tend to have larger parameter counts and computational demands than CNNs \cite{liu2023efficientvit}, and their resource-intensive nature also hinders clinical deployment.

\subsection{Semi-supervised Medical Image Segmentation}

\begin{figure}
    \centering
    \includegraphics[width=0.498\textwidth]{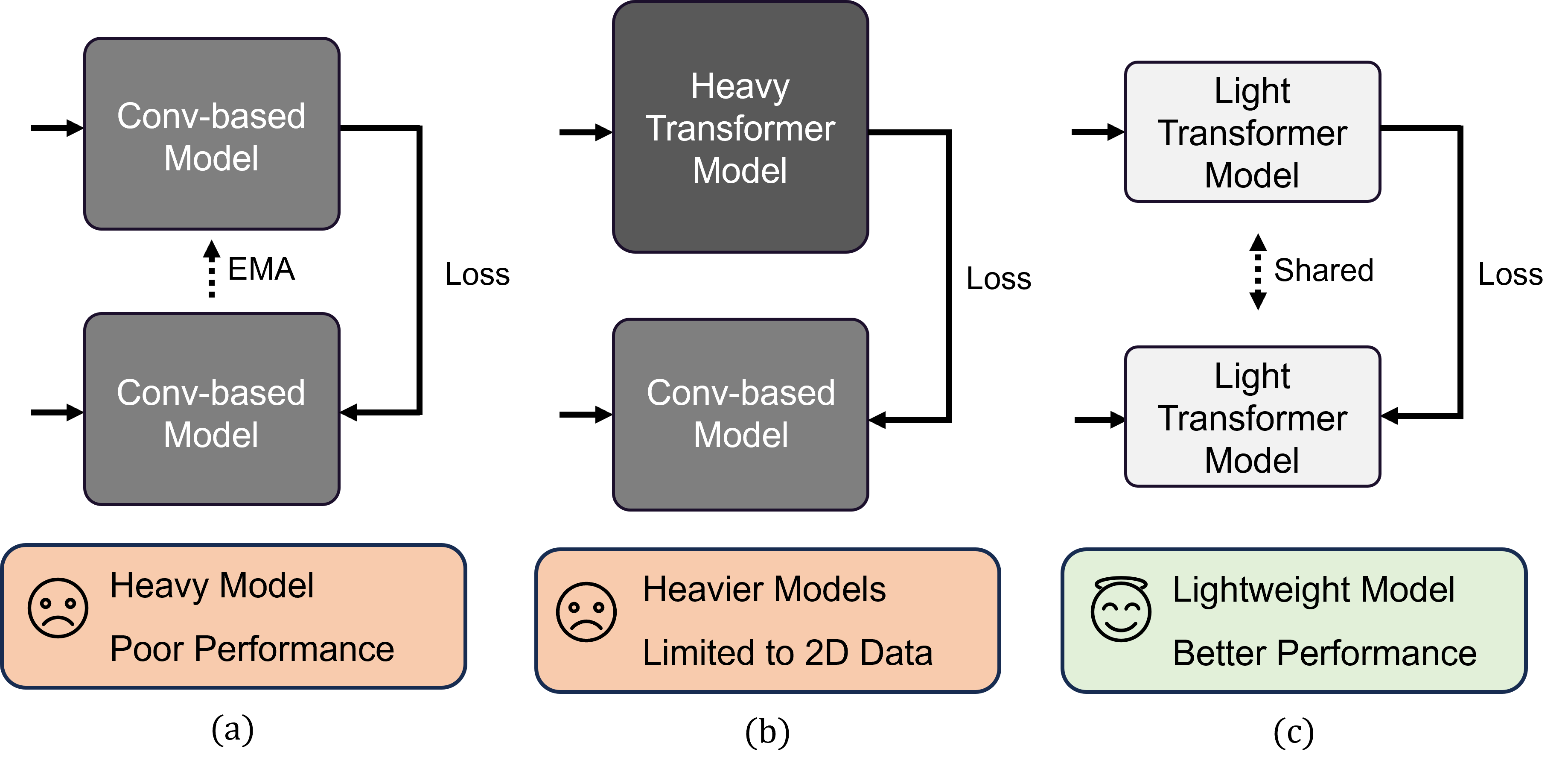}
    \caption{Comparison between different methods for SSMIS. (a) Traditional methods \cite{yu2019uncertaintyawaremeanteacher, wu2022mutual, BCP} combine conv-based medical image segmentation models \cite{vnet} with semi-supervised learning techniques such as mean-teacher \cite{meanteacher}, which are not only computational expensive but also have poor performance. (b) Recent methods \cite{luo2022semicnntransformer} jointly train Transformers \cite{swin} with conv-based models, which are limited to 2D data and rely on natural-domain pretrained models. (c) Our proposed model applies a single shared lightweight Transformer model for SSMIS on 3D medical volumes, which achieves better performance with significantly less computational cost.}
    \label{fig:motivation}
\end{figure}
Recent advances in semi-supervised learning \cite{sohn2020fixmatch, wu2022ssnet, BCP, li2024semantic, zhao2022dcssl, meanteacher, liu2024diffrect, yu2019uncertaintyawaremeanteacher} have significantly mitigated the need for time-consuming manual annotation in medical image segmentation, paving the way for more efficient and accurate image analysis \cite{jiao2022ss4mis_survey}. The core idea of state-of-the-art SSMIS methods is to construct consistency regularization for unlabeled data \cite{sohn2020fixmatch}, which enforces the model to produce consistent predictions for different perturbations and transformations of the same input instance. MC-Net \cite{wu2022mutual} designed a multi-decoder network with mutual consistency constraints to ensure consistency across different decoders. Bai \emph{et al.} \cite{BCP} proposed a method that combines labeled and unlabeled images, using a combination of pseudo-labels from a teacher model and ground truths for supervision. Wu \emph{et al.} \cite{wu2022ssnet} employed a prototype-based strategy to separate the feature manifolds of different classes to produce more compact class-wise distributions. Gao \emph{et al.} \cite{CAML} developed a framework with two models, and regularized them by constraining the omni-correlation matrix of each sub-model to be consistent. Nevertheless, the significant disparity between medical volumes (\textit{e.g.}, MRI \& CT) and natural images hinders the applicability of many transformation techniques \cite{garcea2022data}, which leads to a limited data variety and perturbation design space. Therefore, {these methods may be prone to the test data with distribution shift} \cite{zhao2022dcssl, BCP} and suffer from imprecise segmentation. This motivates us to explore alternative techniques that effectively expand the data diversity \cite{simard2003best} and enhance the model's generalization ability to medical images effectively. Besides, the CNN models traditionally used for SSMIS are often outdated and have a large number of parameters \cite{he2023structured, liu2023generalized}, which makes them less feasible for clinical use, as illustrated in Fig. \ref{fig:motivation}(a). 

While previous methods for SSMIS have primarily relied on CNN-based models such as U-Net or V-Net, recent advances in transformer architecture have sparked interest in exploring their potential in medical image segmentation. Therefore, some studies have attempted to adapt transformers for SSMIS. CTCT \cite{CTCT} designed a co-teaching framework between CNN and Transformer. Huang \emph{et al.} \cite{huang2023semibiuncertain} attached the uncertainty in CNN and transformer levels for alleviating the influence of unreliable pseudo-labels. However, these efforts all require CNNs for joint training to achieve decent performance, as shown in Fig. \ref{fig:motivation}(b). Besides, they all rely on 2D transformers designed for natural images, which are not generalizable to 3D medical data.
Therefore, a pure 3D transformer-based solution for the data-scarce SSMIS task remains a challenging and open problem. In this paper, we address this issue by proposing a light transformer model with an efficient semi-supervised learning strategy as illustrated in Fig. \ref{fig:motivation}(c), which achieves better efficiency and performance than previous works.

\section{Methodology}
\label{sec:methodology}

\begin{figure*}
    \centering
    \includegraphics[width=\textwidth]{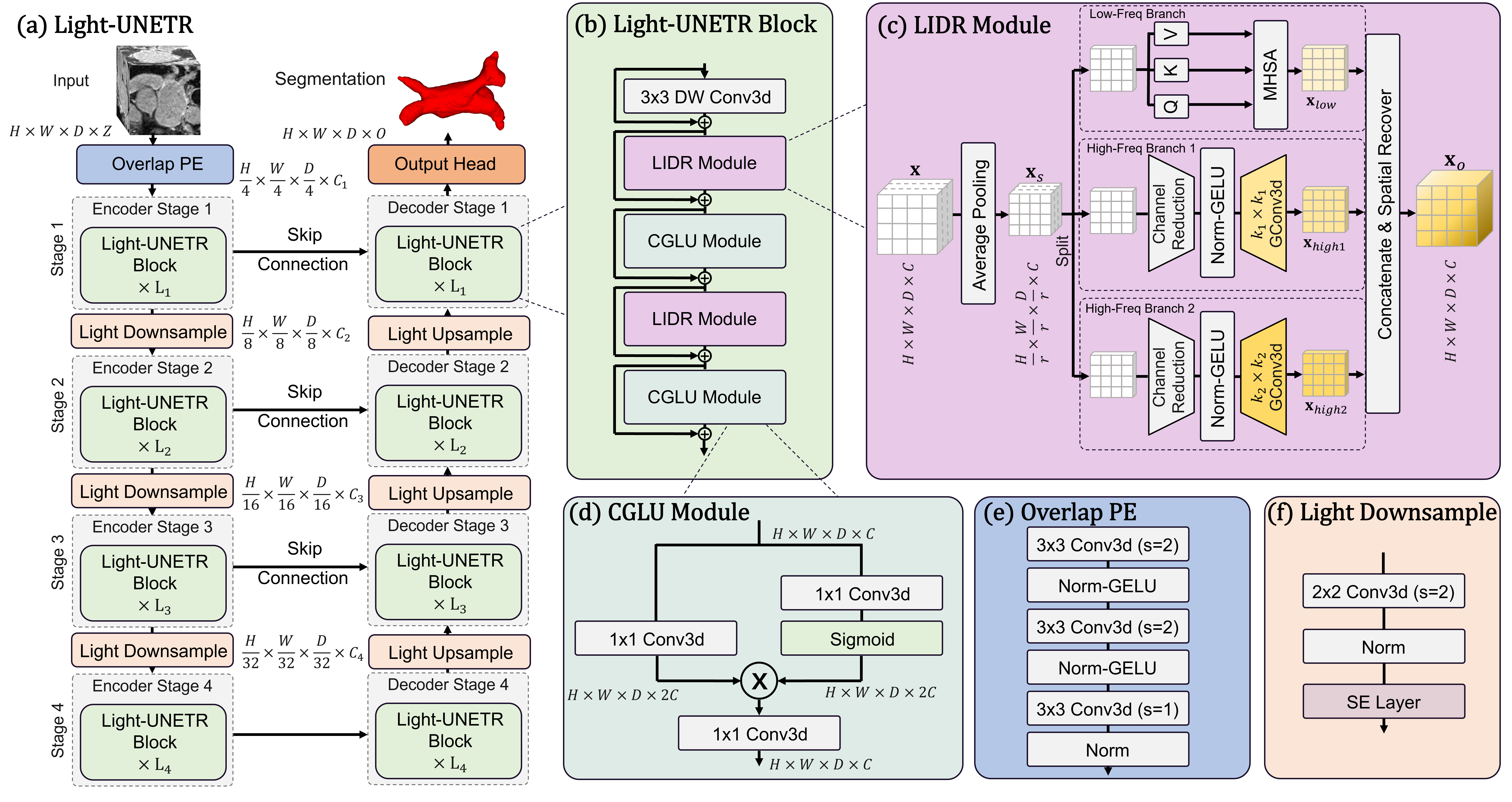}
    \caption{Illustration of the proposed (a) Overall Light-UNETR structure, (b) Light-UNETR Block (c) LIDR module, (d) CGLU module, (e) overlap patch embedding, and (f) Light downsample.}
    \label{fig:enter-label}
\end{figure*}
\subsection{Light-UNETR Architecture}
\subsubsection{Overview}

We first introduce the overall architecture of our Light-UNETR, as presented in Fig. \ref{fig:enter-label}(a), with a UNet-like \cite{unet} encoder-decoder structure with skip connections. Concretely, we introduce overlap patch embedding (Overlap PE) \cite{xiao2021early, pvt_v2} (Fig. \ref{fig:enter-label}(e)) to embed the 3D input volume with $H\times W\times D\times Z$ into tokens with $\frac{H}{4}\times \frac{W}{4}\times \frac{D}{4}\times {C_1}$ dimension, as it could effectively enhance the model capacity in low-level visual representation learning. The architecture comprises four stages in both encoder and decoder, each of which stacks the proposed Light-UNETR building blocks with $C_1, C_2, C_3$ and $C_4$ channels. The spatial dimensions of the feature maps are reduced or expanded by a factor of 2 at each downsampling or upsampling layer, respectively. To achieve efficient downsampling or upsampling, we design a Light Downsample/Upsample layer shown in Fig. \ref{fig:enter-label}(f), which comprises only a convolution or transposed convolution layer, followed by a normalization layer and a squeeze-and-excitation layer \cite{hu2018squeeze}. This design enables the model to selectively emphasize or suppress certain features during resolution changing, thereby improving the representation learning capability and minimizing information loss \cite{zhao2017randomdownsample}. The final output segmentation head upsamples the feature by $4\times$ to recover the original spatial resolution with $O$ number of classes. 

The core of Light-UNETR is the proposed Light-UNTER block, as illustrated in Fig. \ref{fig:enter-label}(b), which is an efficient building block for 3D medical image segmentation. It is composed of a depthwise convolution for local information propagation, followed by $2$ successive lightweight dimension reductive attention (LIDR) and compact gated linear unit (CGLU) modules. The module details are introduced as follows:

\subsubsection{Lightweight Dimension Reductive Attention (LIDR)}
An illustration of LIDR is shown in Fig. \ref{fig:enter-label}(c). Let $\mathbf{x} \in \mathbb{R}^{H \times W \times D \times C }$ be the input tensor, where $C$ is the number of channels, and $H$, $W$, and $D$ are the spatial dimensions. Unlike traditional approaches that directly feed image tokens into multi-head self-attention mechanisms \cite{ViT, deit}, the proposed LIDR Attention module first utilizes an average pooling operator to downsample the input tensor $\mathbf{x}$ with a stride of $r$, resulting in a tensor $\mathbf{x}_s \in \mathbb{R}^{B  \times H/r \times W/r \times D/r \times C}$:
\begin{equation}
\mathbf{x}_s = \text{AvgPool3d}(\mathbf{x}, r).
\end{equation}
Afterwards, the downsampled tensor $\mathbf{x}_s$ is split along the channel dimension into three components with $C/3$ channels each: $[\mathbf{x}_l, \mathbf{x}_{h1}, \mathbf{x}_{h2}] = \text{Split}(\mathbf{x}_s, 3)$, which are then respectively fed into the low-frequency and two high-frequency branches. 
For the low-frequency branch, we employ the MHSA to facilitate information exchange among feature tokens, which aims to capture global structures
and relations in the extracted feature from the medical image.
Thanks to the downsampled input feature with a reduced spatial resolution, we can efficiently compute attention and still capturing global contextual information. Specifically, the self-attention computes query, key, and value vectors via:
\begin{equation}
\mathbf{Q}, \mathbf{K}, \mathbf{V} = \mathbf{W}_q\mathbf{x}_l, \mathbf{W}_k\mathbf{x}_l, \mathbf{W}_v\mathbf{x}_l,
\end{equation}
where $\mathbf{W}_q$, $\mathbf{W}_k$, and $\mathbf{W}_v$ are learnable weights. Next, these matrices are then split into $N_h$ attention heads, where the $h$-th head is computed with the softmax self-attention:
\begin{equation}
\mathbf{x}_h = \text{Softmax}(\mathbf{Q} \mathbf{K}^T )\mathbf{V} = \frac{\exp(\mathbf{Q} \mathbf{K}^T )}{\sum\exp(\mathbf{Q} \mathbf{K}^T )}\mathbf{V}.
\end{equation}
Thus, the output of the low-frequency branch is computed as:
\begin{equation}
\mathbf{x}_{low} = \underset{h \in\left[N_h\right]}{\operatorname{Concat}}(\mathbf{x}_h)
\end{equation}

In addition to the low-frequency information, we employ two high-frequency extractor branches to capture multi-scale fine-grained high frequencies within the input feature. Specifically, the channel dimension of the input feature $\mathbf{x}_{h1/h2}$ is first reduced with $\mathbf{W}_{1/2}$, which efficiently projects the input features into a lower-dimensional space to save computational cost. We then apply batch normalization and a GELU activation function to introduce non-linearity and improve the representational capacity.
Afterwards, we apply group convolution to simultaneously capture high-frequency changes such as local edges or textures, meanwhile restore the original channel dimension with minimal computational overhead. Specially, we apply two different kernel sizes $k_1$ and $k_2$ for the two branches. By using different kernel sizes, each branch can extract features at different spatial resolutions, enabling the model to learn representations that are more invariant to scale variations. 
Formally, the operations performed by the two branches can be expressed as:
\begin{equation}
\begin{aligned}
        \mathbf{x}_{high1} &= \text{GConv}(\text{GELU}(\text{BN}(\mathbf{W}_1(\mathbf{x}_{h1}))), k_1), \\
    \mathbf{x}_{high2} &= \text{GConv}(\text{GELU}(\text{BN}(\mathbf{W}_2(\mathbf{x}_{h2}))), k_2),
\end{aligned}
\end{equation}
Finally, the output of the LIDR module is computed by concatenating the low-frequency feature $\mathbf{x}_{{low}}$ and the two high-frequency features $\mathbf{x}_{{high1}}$ and $\mathbf{x}_{{high2}}$, and the spatial resolution is recovered via a transposed depthwise convolution with a stride of $r$:
\begin{equation}
\mathbf{x}_{{o}} = \text{ConvTranspose}(\text{Concat}(\mathbf{x}_{{low}}, \mathbf{x}_{{high1}}, \mathbf{x}_{{high2}}), r).
\end{equation}
With separate branches to capture low and high frequency features, the model can better disentangle and encode different aspects of the input, and is encouraged to learn more diverse representations in a computational efficient manner.

\subsubsection{Compact Gated Linear Unit (CGLU)}
Existing works have demonstrated that the channel communication is critical for lightweight ViTs \cite{liu2023efficientvit}. However, the Feed-Forward Networks (FFNs) have relatively large number of parameters, which can lead to overfitting of lightweight models and increased computational complexity. The Gated Linear Unit (GLU) has been demonstrated to outperform FFNs as a powerful channel mixer in natural language processing \cite{shazeer2020glu}, yet its potential in medical imaging is still unexplored. Therefore, we propose a Compact Gated Linear Unit (CGLU) that enables the model to selectively focus on particular channels efficiently, allowing for better feature extraction and interaction in medical images. We illustrate the operation of CGLU in Fig. \ref{fig:enter-label}(d). Specifically for input feature $\mathbf{x}$, we have:
\begin{equation}
\begin{aligned}
\mathbf{x}_{emb} &= \mathbf{W}_\text{id}\mathbf{x} \otimes (\sigma(\mathbf{W}_\text{gate} \mathbf{x})), \\
\mathbf{x}_{cglu} &= \mathbf{W}_\text{rec}\mathbf{x}_{emb},
\end{aligned}
\end{equation}
where $\mathbf{W}_\text{id}$ and $\mathbf{W}_\text{gate}$ are linear layers that perform identity transformation and gating operation, both with shape $\mathbb{R}^{2C\times C}$. $\mathbf{W}_\text{rec}$ is a dimension recovery layer with shape $\mathbb{R}^{C\times 2C}$, $\sigma$ is the Sigmoid function, and $\otimes$ denotes the element-wise multiplication. Notably, the CGLU uses a reduced expansion factor of 2, which is designed to be more parameter-efficient than the original GLU \cite{shazeer2020glu} that has a larger expansion factor of $8/3$. This allows the model to capture complex patterns and relationships in medical images, while minimizing the number of parameters and computations with the compact design.

\subsubsection{\xy{Large Variant of Light-UNETR Model}}
\xy{To fairly compare with recent high–capacity and efficient 3D segmentation
frameworks such as nnUNet~\cite{isensee2021nnunet} and MedNeXt~\cite{roy2023mednext},
we additionally design a \emph{large} variant called {Light-UNETR-L}.
It follows the popular $5$ stage encoder–decoder macro architecture that is commonly adopted by state-of-the-art methods yet preserves the lightweight design of Light-UNETR blocks. Specially, Light-UNETR-L contains a total of five resolution levels $1\times, 2\times, 4\times, 8\times, 16\times$. Instead of using the overlap patch embedding, the first two stages, upsample, and downsample are replaced with residual convolutional blocks similar to \cite{roy2023mednext}, which aims to provide an early and inexpensive {feature funnel}. From the third stage onward, we employ our proposed Light-UNETR block exactly as in the original model. The standard Light-UNETR demonstrates that our blocks can obtain strong accuracy at minimal cost. Light-UNETR-L shows that the same design scales gracefully to a higher capacity regime, allowing direct and controlled comparison to heavyweight models without changing the computational paradigm.}

\subsection{Contextual Synergic Enhancement}

\begin{figure}
    \centering
    \includegraphics[width=0.49\textwidth]{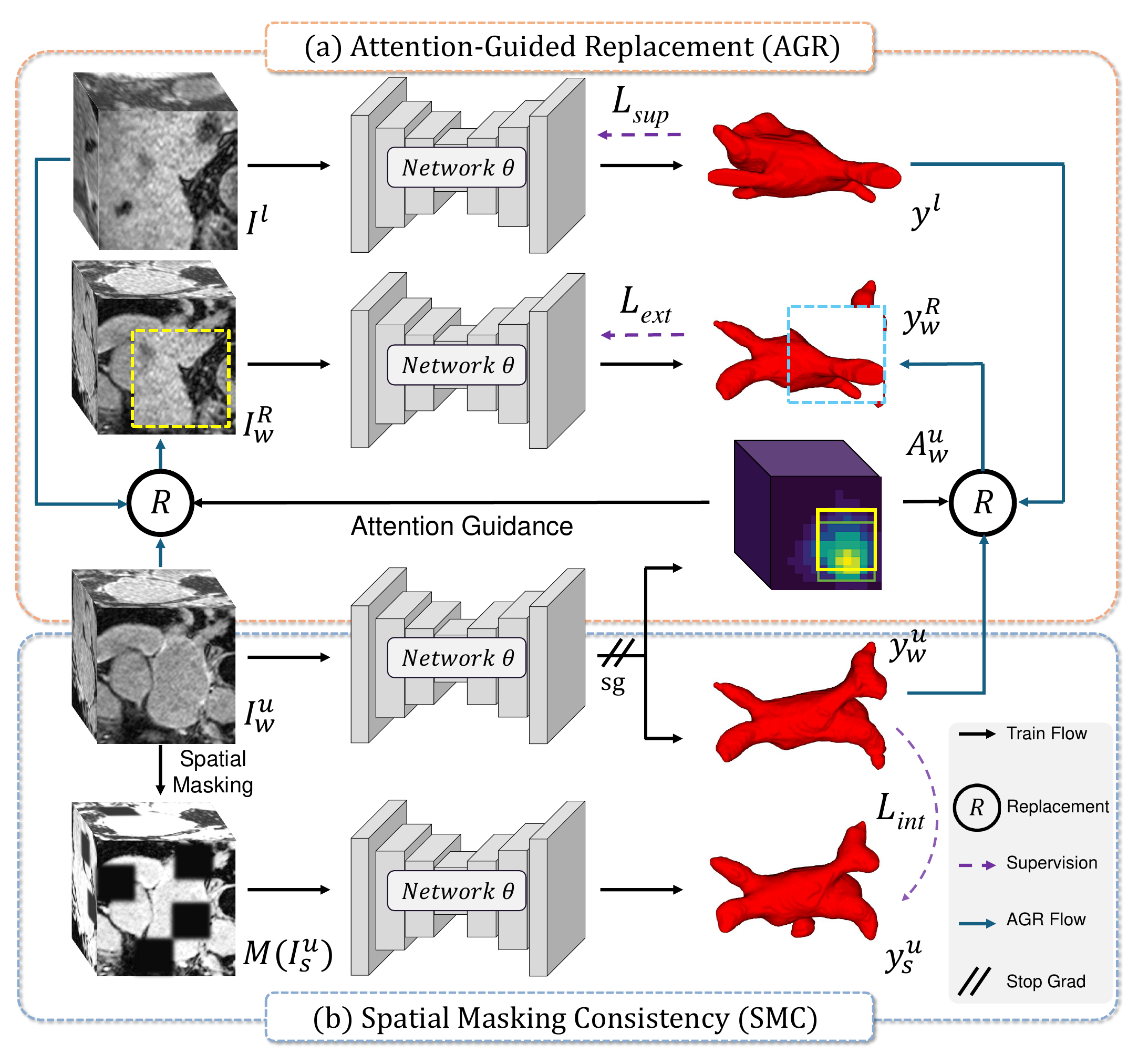}
    \caption{Illustration of the CSE semi-supervised learning framework, which contains (a) Attention-Guided Replacement (AGR) and (b) Spatial Masking Consistency (SMC).}
    \label{fig:CSE}
\end{figure}
To unlock the full potential of contextual information for SSMIS, we introduce the CSE semi-supervised learning paradigm, as shown in Fig. \ref{fig:CSE}. It comprises two synergistic components: an Attention-Guided Replacement strategy that leverages extrinsic context to infer semantic meaning in unlabeled data, and a Spatial Masking Consistency strategy that facilitates intrinsic context reasoning. Specifically, it utilizes a single Light-UNETR network parameterized with $\theta$ to process labeled image $I^l$ and unlabeled image $I^u$. The unlabeled image $I^u$ is augmented with weak and strong operators, resulting in two different versions $I^u_w$ and $I^u_s$. Here we introduce the proposed strategies in detail.

\subsubsection{Attention-Guided Replacement for Extrinsic Contextual Information}

We propose an Attention-Guided Replacement data perturbation method which selectively mixes regions of the unlabeled image with labeled image based on attention values, as shown in Fig. \ref{fig:CSE}(a). The proposed method is designed to effectively utilize the extrinsic contextual information in labeled data to support the learning of unlabeled data, thereby achieving a mutually beneficial effect, while the attention maps serve as appropriate importance guidance for saliency regions \cite{ren2023sg, choi2022tokenmixup}. Specifically, let $I_w^u \in \mathbb{R}^{H \times W \times D \times Z}$ denoting the weakly augmented input unlabeled image with $H$, $W$, $D$, and $Z$ as the height, width, depth, and input channels, we can obtain the pseudo label and the final-layer attention map via the segmentation network $\theta$:
\begin{equation}
    \mathbf{A}, y^u_w = F(I^u_w; \theta),
\end{equation}
where $y^u_w \in \mathbb{R}^{H \times W \times D \times O}$ and $\mathbf{A} \in \mathbb{R}^{H \times W \times D \times 1}$. To utilize the attention map tensor $\mathbf{A}$ to guide the masking process,
\xy{we split the input image into $N$ patch regions $\{R_i\}_{i=1}^N$ of size $P \times P \times P$, where $P = \lfloor H \times \alpha \rfloor$, and $\alpha$ is a hyperparameter controlling the patch size. We parameterize the regions with regard to their coordinates as $(x, y, z) \in \{0, 1, \dots, N-1\}^3$, where $N = \lceil H / P \rceil$. For each patch, we calculate the attention value by summing the attention map values within that patch:}

\xy{\begin{equation}
a_{x, y, z} = \sum_{i=x}^{x+P-1} \sum_{j=y}^{y+P-1} \sum_{k=z}^{z+P-1} \mathbf{A}[:, i, j, k],
\end{equation}
we then normalize the attention values to probabilities using the softmax function:}

\xy{\begin{equation}
p_{x, y, z} = \frac{\exp(a_{x, y, z})}{\sum_{x'=0}^{N-1} \sum_{y'=0}^{N-1} \sum_{z'=0}^{N-1} \exp(a_{x', y', z'})}.
\end{equation}}

\noindent\xy{To adaptively select the regions that have larger attention values while maintain stochasticity, we sample the normalized attention values using the probability distribution $\mathbf{P}(x, y, z) = p_{x, y, z}$ and obtain the indexed coordinates of the region $\{\hat{x}, \hat{y}, \hat{z}\} = \text{Index}(\mathbf{P}(x, y, z))$.
and we perform a replacement on the selected patch from a sampled labeled image $I^l$:
\begin{equation}
I^R_w = R(I_w^u, I^l, [\hat{x}:\hat{x}+P, \hat{y}:\hat{y}+P, \hat{z}:\hat{z}+P]),
\end{equation}
where $R(\cdot)$ is a replacement function that replaces the values within the region $[\hat{x}:\hat{x}+P, \hat{y}:\hat{y}+P, \hat{z}:\hat{z}+P]$ in the input image $I_w^u$ with the corresponding values from the labeled image $I^l$, and leading to a mixed data $I^R_w$. Similarly, the replacement is also applied on the pseudo-label to generate a mixed label $y^R_w$. The full process is summarized in Alg. \ref{alg:AGR}.}
Our method effectively incorporates the label information from the labeled image $I^l$ into the input image $I_w^u$, allowing the model to adaptively use confident extrinsic contextual information to guide the learning of less confident regions, which is crucial for reasoning the ambiguity in the unlabeled data.

\begin{algorithm}[t]
\caption{\xy{Attention-Guided Replacement}}
\begin{algorithmic}[1]
\Require Unlabeled image $I_w^u$; Labeled image $I^l$ and its label $y^l$; Segmentation network $F$ parameterized with $\theta$
\State Obtain $(y_w^u, \mathbf{A}) \gets F(I_w^u;\theta)$ 
\State Divide $\mathbf{A}$ into $N$ patched regions $\{R_i\}_{i=1}^N$, each region $R_i$ is represented by patch coordinates $(x,y,z)$
\For{each $R_i$}
    \State Compute attention value $a_i = \sum \mathbf{A}$ over region $R_i$
\EndFor
\State Normalize $\{a_i\}$ across all patches as probabilities: $p_i = a_i / \sum_j a_j$
\State Sample the $k$-th patch region $R_k$ according to $\{p_i\}$
\State Replace $R_k$ in $I_w^u$ with the patch at the same location in $I^l$ to obtain $I_w^R$
\State Replace $R_k$ in $y_w^u$ with the patch at the same location in $y^l$ to obtain $y_w^R$
\end{algorithmic}
\label{alg:AGR}
\end{algorithm}

\subsubsection{Spatial Masking Consistency for Intrinsic Contextual Information}

The incorporation of intrinsic contextual information is essential for accurately segmenting medical images, particularly when dealing with ambiguous regions. This is because the human body exhibits a characteristic structural organization \cite{zhao2021diagnose}, which can be leveraged to inform anatomical prior knowledge \cite{paragios2003level}. For instance, the spatial relationships between specific organs and surrounding tissues can provide valuable cues for determining their boundaries, even when they are not clearly defined. However, how to leverage the intrinsic spatial context for improving SSMIS remains unexplored.

We present Spatial Masking Consistency to address the aforementioned limitation, which is shown in Fig. \ref{fig:CSE}(b). It leverages the weakly and strongly augmented versions of the unlabeled data $I^u$, which are ${I}_w^u$ and $I_s^u$. Subsequently, we define the parameters for the spatial masking template $\mathcal{M}$, characterized by a 3D mask region side length of $l$ and a mask ratio of $v$. To implement the masking, we first resize an all-ones volume to $\mathcal{M}'$ with dimensions $H/l \times W/l \times D/l \times 1$, matching the size of the input data. We then randomly select a proportion of $v$ voxels and set their corresponding values to 0, effectively masking them out. Next, we upsample $\mathcal{M}'$ to the original size using trilinear interpolation, resulting in the mask $\mathcal{M}\in \mathbb{R}^{H\times W\times D \times 1}$. Finally, we apply the mask to $I_s^u$ by performing a Hadamard product between the two,
\begin{equation}
    \mathcal{M}(I_s^u) = \mathcal{M} \cdot \mathcal{V}_s,
\end{equation}
where $\mathcal{M}(I_s^u)$ refers to the strong views after spatial masking. Afterwards, the framework enforces the consistency between the segmentation maps generated from the original data and the masked version data, which encourages the model to infer ambiguous regions via the spatial contextual information. Compared to the previous masking methods \cite{hoyer2023mic, gao2023masked, xie2022simmim, devries2017cutout}, the proposed masking has a smooth edge, which prohibits the model from learning the abrupt transitions, meanwhile boosts the model's capacity in learning intrinsic contextual information.  

\subsection{Optimization}

The overall optimization objective of CSE comprises three components. Firstly, the supervised loss is applied via a dice loss between the labeled data and its corresponding label:
\begin{equation}
    L_{sup} = Dice(F(I^l; \theta), y^l),
\end{equation}
where the dice loss is computed as:
\begin{equation}
   Dice(R, Q) = 1 - \frac{2 \times \left| R \cap Q \right|}{\left| R \right| + \left| Q \right|}.
\end{equation}
Secondly, we apply the loss for learning extrinsic context $L_{ext}$ with the replaced data $I^R_w$:
\begin{equation}
    L_{ext} = Dice(F(I^R_w; \theta), y^R_w).
\end{equation}
Lastly, the $L_{int}$ is employed on the model to learn intrinsic context with the masked data:
\begin{equation}
    L_{int} = Dice(F(\mathcal{M}(I_s^u); \theta), y^u_w).
\end{equation}
Finally, the total loss of the proposed network is defined as follows with a linear combination:
\begin{equation}
    L = L_{sup} + \lambda_1 L_{ext} + \lambda_2 L_{int},
\label{eq:final_loss}
\end{equation}
where $\lambda_1$ and $\lambda_2$ are balancing factors that are trade-off weights to adjust the
importance of loss components, which are empirically set to 4 and 1, respectively. Our model is trained in an end-to-end manner. Through optimizing with Eq. (\ref{eq:final_loss}), our SSMIS framework can obtain the remarkable performance. During inference, the testing data is directly fed into the well-trained segmentation network to generate the predicted segmentation map.

\section{Experiments}
\label{sec:experiments}

\subsection{Datasets}
To evaluate the effectiveness of the proposed framework, we conduct experiments on three widely recognized SSMIS datasets and compare with other prior works:

\textbf{LA.} The Atrial Segmentation Challenge \cite{LA} dataset includes 100 3D gadolinium-enhanced magnetic resonance image scans (GE-MRIs), and are annotated with 3D binary masks of the left atrial (LA) cavity. Following common practice \cite{yu2019uncertaintyawaremeanteacher, BCP}, we utilize 80 volumes as the training dataset and the remaining 20 volumes are used as the test dataset.

\textbf{Pancreas-CT.} The Pancreas-CT \cite{pancreas} dataset was collected by the National Institutes of Health Clinical Center. It contains 82 abdominal 3D CT volumes for pancreas segmentation. In this work, we follow \cite{xiang2022fussnet, BCP, luo2022semipyramid} to split 62 volumes for training and 20 remaining volumes for testing.

\textbf{BraTS 2019.} The BraTS 2019 \cite{Brats} dataset is from the Multimodal Brain Tumor Segmentation Challenge 2019, comprising 335 volumes with four modalities: FLAIR, T1, T1ce and T2. The task is to segment the whole tumors from FLAIR images. The scans are split into 250, 25 and 60 volumes for training, validation, and testing.

To evaluate the generalizability of our method, we also perform additional benchmarking on three fully-supervised segmentation tasks spanning different anatomical structures: MSD Task01 (brain tumor) \cite{decathlon}, AbdomenCT-1K (multi-organ abdomen) \cite{ma2021abdomenct}, and HNC Tumor dataset \cite{HNCtumor}.

\subsection{Experiment Setup and Evaluation Metrics}
We conduct all experiments on NVIDIA 4090 GPUs with PyTorch version 2.4.1 and CUDA version 12.2. For semi-supervised experiments, all methods are trained for 15k iterations with batch size 4, using the SGD optimizer with initial learning rate 0.01 and a cosine scheduler with 500 warm-up iterations. Each batch contains 50\% labeled data and 50\% unlabeled data. Random cropping is used as the basic weak augmentations $\mathcal{A}_w$, and we apply random gamma adjustment \cite{simclrv2} as the specific strong augmentation in $\mathcal{A}_s$. During training, the scans are cropped to $112\times112\times80$ in LA, $96\times96\times96$ in Pancreas-CT and BraTS 2019,
respectively, which strictly follow the settings in \cite{BCP, xiang2022fussnet, su2024mutual_MLRP}. Threshold $\tau$ for pseudo-labeling is fixed to 0.75. The configuration of the model includes $\{C_1, C_2, C_3, C_4\}=\{24, 48, 60, 96\}$ and $\{L_1, L_2, L_3, L_4\}=\{1, 2, 3, 2\}$. The spatial reduction ratio $r$ in LIDR is set to $\{4,2,2,1\}$ for the four stages. $k_1$ and $k_2$ are set to 3 and 5, respectively.
During inference, a sliding window is used to obtain segmentation results for 3D datasets following common practice \cite{BCP, xiang2022fussnet}, with a stride of $18\times18\times4$ on LA, $16\times16\times4$ on Pancreas-CT, and $64\times64\times64$ on BraTS 2019. 
We report four metrics that are commonly used in medical image segmentation \cite{wu2022mutual, BCP, CAML}: Dice Coefficient (\textit{Dice}), Jaccard Score (\textit{Jac}), 95\% Hausdorff Distance (\textit{95HD}), and Average Surface Distance (\textit{ASD}). \xy{Moreover, to evaluate the effectiveness our proposed lightweight architecture design, we additionally conducted fully-supervised segmentation experiments on MSD-Task01 \cite{decathlon}, AbdomenCT-1K \cite{ma2021abdomenct}, and HNC Tumor \cite{HNCtumor}. We train the models with an initial learning rate 0.0001, optimizer AdamW and batch size 1. The channel configuration of each stage in Light-UNETR-L is set to \{32, 60, 96, 150, 198\}. The per-class Dice coefficient is reported and compared.}

\begin{table*}[t]
  \centering
  \caption{
\label{tab:merged-results}Semi-supervised segmentation performance on LA and Pancreas-CT datasets. Bold denotes the best result. $\uparrow$ and $\downarrow$ denote that higher or lower values are better, respectively. L/U refers to the number of labeled/unlabeled scans. All models use the proposed Light-UNETR as base segmentation model.}
\setlength{\tabcolsep}{1.4mm}
\begin{tabular}{c|c|cccc|c|cccc}
\toprule
\multirow{3}{*}{Method}   & \multicolumn{5}{c|}{{LA dataset}}   & \multicolumn{5}{c}{{Pancreas-CT dataset}} \\
\cmidrule(lr){2-6} \cmidrule(lr){7-11}
& L/U & Dice (\%)$\uparrow$ & Jac (\%)$\uparrow$ & 95HD (mm)$\downarrow$ & ASD (mm)$\downarrow$ & L/U & Dice (\%)$\uparrow$ & Jac (\%)$\uparrow$ & 95HD (mm)$\downarrow$ & ASD (mm)$\downarrow$ \\

\midrule
Light-UNETR (Ours)             & 4/0    & 75.55   & 61.99   & 17.80   & 6.16   & 6/0     & 59.46   & 44.15   & 22.64   & 7.92 \\
UA-MT (MICCAI'19)  \cite{yu2019uncertaintyawaremeanteacher}   & 4/76    & 78.34   & 65.11   & 15.42   & 4.69   & 6/56    & 52.41   & 36.80   & 36.34   & 11.78 \\
SS-Net (MICCAI'22)  \cite{wu2022ssnet}                        & 4/76    & 79.69   & 67.02   & 12.44   & 3.86   & 6/56    & 51.31   & 36.10   & 22.04   & 7.50 \\
CAML (MICCAI'23)  \cite{CAML}                                 & 4/76    & 79.74   & 66.94   & 14.58   & 4.46   & 6/56    & 57.34   & 40.94   & 28.41   & 8.86 \\
BCP (CVPR'23)  \cite{BCP}                                     & 4/76    & 84.76   & 73.88   & 10.22   & 2.89   & 6/56    & 68.93   & 53.86   & 14.01   & 4.37 \\
MLRP (MedIA'24)  \cite{su2024mutual_MLRP}                     & 4/76    & 81.27   & 69.36   & 10.70   & 3.12   & 6/56    & 67.05   & 51.44   & 13.71   & 4.39 \\
\textbf{CSE-Light-UNETR (Ours)}          & 4/76    & \textbf{89.53}  & \textbf{81.15}  & \textbf{6.14}   & \textbf{1.88}  & 6/56     & \textbf{73.77}  & \textbf{59.44}  & \textbf{11.77}  & \textbf{3.76} \\
\midrule
Light-UNETR (Ours)             & 8/0     & 80.05   & 69.86   & 12.88   & 4.13   & 12/0    & 67.28   & 51.99   & 14.73   & 4.12 \\
UA-MT (MICCAI'19)  \cite{yu2019uncertaintyawaremeanteacher}   & 8/72    & 82.53   & 70.73   & 11.80   & 3.84   & 12/50   & 67.60   & 52.23   & 18.06   & 6.39 \\
SS-Net (MICCAI'22) \cite{wu2022ssnet}                         & 8/72    & 83.51   & 72.19   & 9.93    & 2.78   & 12/50   & 68.85   & 54.10   & 14.44   & 4.68 \\
CAML (MICCAI'23) \cite{CAML}                                  & 8/72    & 82.36   & 70.56   & 9.79    & 3.08   & 12/50   & 68.76   & 52.86   & 11.96   & 3.92 \\
BCP (CVPR'23) \cite{BCP}                                      & 8/72    & 86.81   & 76.94   & 8.91    & 2.49   & 12/50   & 73.15   & 58.56   & 8.61    & 2.41 \\
MLRP (MedIA'24)   \cite{su2024mutual_MLRP}                    & 8/72    & 82.92   & 71.18   & 10.24   & 2.96   & 12/50   & 71.80   & 57.24   & 8.84    & 2.45 \\
\textbf{CSE-Light-UNETR (Ours)}           & 8/72    & \textbf{90.50}  & \textbf{82.74}  & \textbf{5.82}   & \textbf{1.83}  & 12/50    & \textbf{78.50}  & \textbf{65.27}  & \textbf{8.35}   & \textbf{2.24} \\
\midrule
{Fully Supervised} & {80/0} & {91.58} & {84.51} & {4.95} & {1.43} & {62/0} & 80.45 & 67.96 & 6.19 & 1.30 \\
\bottomrule
\end{tabular}
\end{table*}

\begin{table}[t]
  \centering
  \caption{Semi-supervised segmentation performance on the BraTS 2019 dataset. Bold denotes the best result. $\uparrow$ and $\downarrow$ denote that higher or lower results are preferred. L/U refers to the number of labeled/unlabeled scans. All models use the proposed Light-UNETR as base segmentation model.}
\setlength{\tabcolsep}{1.6mm}{\begin{tabular}[width=\textwidth]{c|c|cccc}
\toprule
\multirow{2}{*}{Methods} & \multicolumn{5}{c}{BraTS 2019 dataset}                 \\
\cmidrule{2-6}
                         & L/U     & Dice$\uparrow$ & Jac$\uparrow$ & 95HD$\downarrow$ & ASD$\downarrow$ \\
\midrule
Light-UNETR (Ours)                 & 25/0     &   75.91 & 64.80 & 12.93 & 3.86 \\
UA-MT (MICCAI'19)  \cite{yu2019uncertaintyawaremeanteacher}                  & 25/225    & 74.13 & 62.45 & 16.23 &  4.42  \\
SS-Net  (MICCAI'22)           \cite{wu2022ssnet}       & 25/225    &  78.79& 67.96& 13.65&  3.96        \\
CAML  (MICCAI'23)   \cite{CAML}                  & 25/225   &  78.35 &  67.02 &13.34&  3.85   \\
BCP (CVPR'23)  \cite{BCP}                      & 25/225   & 77.06&   66.02& 18.10&  5.33   \\
MLRP (MedIA'24)  \cite{su2024mutual_MLRP}& 25/225 & 78.10 &  67.51&12.50&  3.79   \\
\textbf{CSE-Light-UNETR (Ours)}          & 25/225    &  \textbf{79.73}& \textbf{68.76}& \textbf{11.65}&\textbf{2.25} \\
\midrule
{Fully Supervised} & {250/0} & 79.93 & 69.33 & 11.44 & 2.26  \\
\bottomrule
\end{tabular}}
\label{tab:results-brats}
\end{table}

\subsection{Semi-supervised Medical Image Segmentation}
We compare our proposed CSE-Light-UNETR with state-of-the-art SSMIS methods UA-MT \cite{yu2019uncertaintyawaremeanteacher}, SS-Net \cite{wu2022ssnet}, CAML \cite{CAML}, BCP \cite{BCP}, and MLRP \cite{su2024mutual_MLRP}. All methods are trained with the original configurations as presented in their released code with the proposed Light-UNETR as the base segmentation model. Moreover, we display the results of the fully-supervised Light-UNETR model with labeled data only to clearly demonstrate the improvements of each method. The detailed results on each dataset are given below:
\subsubsection{Results on LA dataset}
\begin{figure*}[t]
    \centering
    \hspace{-0.5cm} 
    \includegraphics[width=0.88\textwidth]{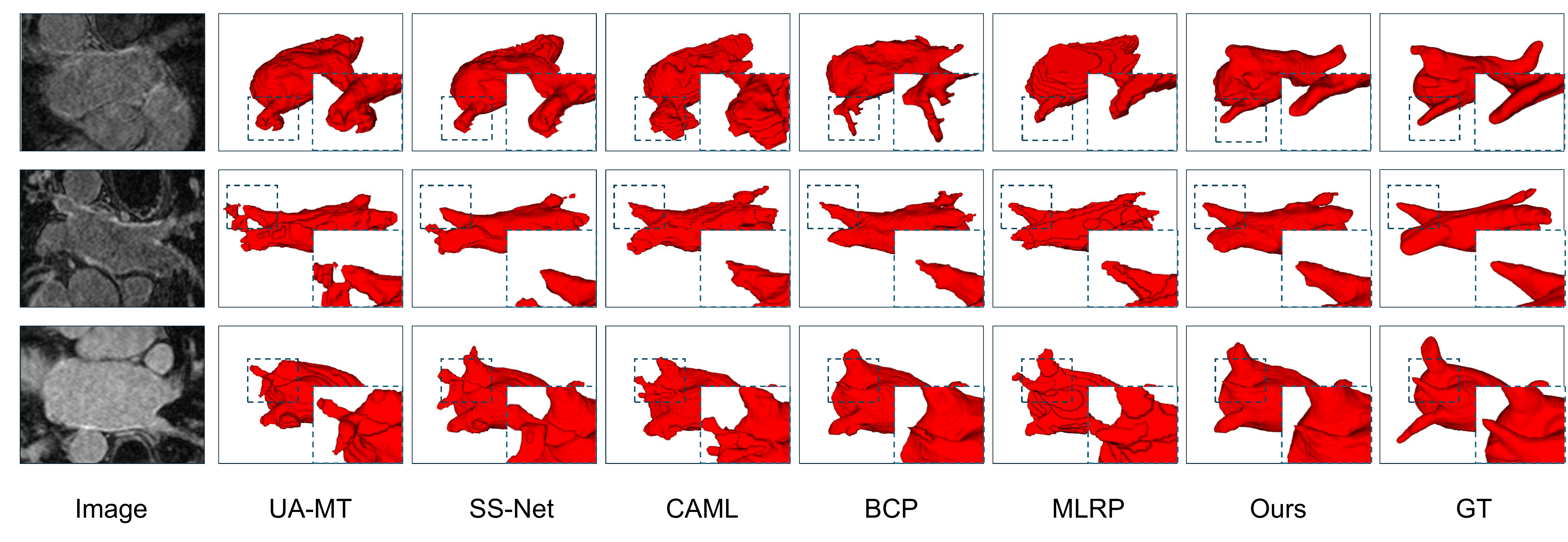}
    \caption{\xy{Qualitative segmentation results comparison of different methods on the LA dataset with 5\% labeled data. From left-to-right are UA-MT \cite{yu2019uncertaintyawaremeanteacher}, SS-Net \cite{wu2022ssnet}, CAML \cite{CAML}, BCP \cite{BCP}, MLRP \cite{su2024mutual_MLRP}, our CSE-Light-UNETR, and ground-truth.}}
    \label{fig:quali:la}
\end{figure*}

The segmentation performance of various methods on the LA dataset is presented in the Tab. \ref{tab:merged-results}. With only 4 (5\%) and 8 (10\%) labeled scans, our Light-UNETR model demonstrates a strong baseline for SSMIS, achieving Dice scores of 75.55\% and 80.05\%, respectively. When augmented with the CSE framework, which incorporates both labeled and unlabeled data, our CSE-Light-UNETR model significantly outperforms the baseline. On the 4 labeled scenario, the model improves to a Dice score of 89.53\% and a Jaccard score of 81.15\%, showcasing the effectiveness of CSE. When comparing with BCP \cite{BCP}, our method surpasses it by 4.77\% Dice and 7.27\% Jaccard, demonstrating its ability in leveraging contextual information for SSL. In the scenario with 8 labeled scans, our model further enhances its performance, achieving a Dice score of 90.50\%. Comparing to MLRP \cite{su2024mutual_MLRP} with a Dice score of 82.92\%, our model outperforms it by 7.58\%, highlighting the effectiveness of our model in using both labeled and unlabeled data to achieve accurate segmentation results. We illustrate the predicted segmentation masks of different methods in Fig. \ref{fig:quali:la}. The proposed CSE-Light-UNETR can generate more precise boundaries for the left atrial (2$^{nd}$ \& 3 $^{rd}$) rows and can effectively capture the semantics of the anatomy and infer hard regions (1$^{st}$ row). The qualitative results further validate the importance of contextual learning for unlabeled medical data.

\subsubsection{Results on Pancreas-CT dataset}

\begin{figure*}[t]
    \centering
    \includegraphics[width=0.88\textwidth]{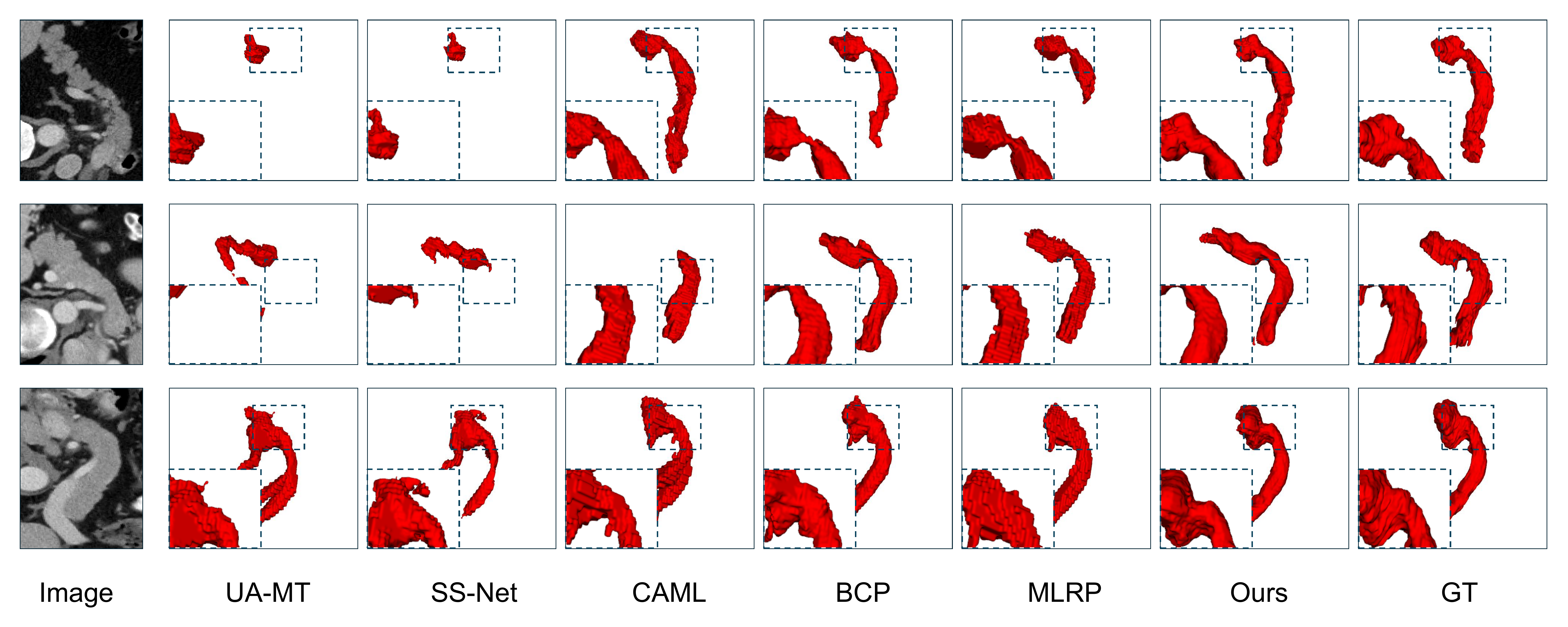}
    \caption{\xy{Qualitative segmentation results comparison of different methods on the Pancreas-CT dataset with 10\% labeled data. From left-to-right are UA-MT \cite{yu2019uncertaintyawaremeanteacher}, SS-Net \cite{wu2022ssnet}, CAML \cite{CAML}, BCP \cite{BCP}, MLRP \cite{su2024mutual_MLRP}, our CSE-Light-UNETR, and ground-truth.}}
    \label{fig:quali:pancreasct}
\end{figure*}

\begin{table*}[h!]
\centering
\caption{\label{fullysupervised_part1}\xyl{Fully-supervised segmentation results on LA, Pancreas and BraTS 2019 datasets. The best results are highlighted in \textbf{bold}, and the second-best are \underline{underlined}. FLOPs are computed with spatial size (112, 112, 80).}}
\resizebox{\textwidth}{!}{
\begin{tabular}{l|cccc|cccc|cccc|cc}
\toprule
\multirow{3}{*}{{Method}} 
& \multicolumn{4}{c|}{{LA dataset}} 
& \multicolumn{4}{c|}{{Pancreas-CT dataset}} 
& \multicolumn{4}{c|}{{BraTS 2019 dataset}} 
& \multirow{3}{*}{{FLOPs (G)}} 
& \multirow{3}{*}{{Params (M)}}  \\
\cmidrule(lr){2-5} \cmidrule(lr){6-9} \cmidrule(lr){10-13}
& {Dice$\uparrow$} & {Jac$\uparrow$} & {95HD$\downarrow$} & {ASD$\downarrow$}  
& {Dice$\uparrow$} & {Jac$\uparrow$} & {95HD$\downarrow$} & {ASD$\downarrow$}  
& {Dice$\uparrow$} & {Jac$\uparrow$} & {95HD$\downarrow$} & {ASD$\downarrow$}  
& &
\\ 
\midrule
\multicolumn{15}{c}{\textit{Foundation Models}}\\
\midrule
MedSAM2 \cite{ma2025medsam2}            & 80.44 & 67.67 & 21.59 & 5.28       & 69.01 & 53.22 & 22.23 & 5.31       & 83.08 & 74.02 & 17.21 & 6.36      & 2,259.2 & 38.96 \\
Efficient-MedSAM2 \cite{ma2025medsam2}  & 78.61 & 65.39 & 21.82 & 6.02       & 68.92 & 53.06 & 18.83 & 5.55       & 82.29 & 71.06 & 11.13 & 6.01      & 1,922.0 & 34.10\\
Med-FastSAM  \cite{luomedfastsam}       & 83.16 & 69.53 & 19.02 & 5.10       & 70.30 & 59.71 & 19.72 & 4.88       & 82.06 & 71.58 & 10.84 & 5.03      & 1,784.9 & 14.48  \\
\midrule
\multicolumn{15}{c}{\textit{Segmentation Networks}}\\
\midrule
nnUNet   \cite{isensee2021nnunet}           & 70.62 & 63.93 & 25.45 & 1.73       & 76.76 & 64.21 & 16.03 & {1.14}      & 77.29 & 67.40 & 12.28 & 2.09      & 227.88 & 13.82 \\
SegResNet \cite{segresnet}          & 84.40 & 76.45 & 10.50 & 1.93       & 81.37 & 69.21 & 7.76  & 1.21       & 77.67 & 67.17 & 12.13 & 2.02      & 229.87 & 10.64 \\
TransBTS \cite{wang2021transbts}           & 69.76 & 59.79 & 23.45 & 4.61       & 76.43 & 63.09 & 11.82 & 1.62       & 79.24 & 69.42 & \underline{9.61} & 1.79     & 161.14 & 30.61 \\
Swin UNETR    \cite{he2023swinunetr}      & 87.86 & 79.54 & 7.08  & 2.00       & 78.83 & 66.13 & 9.96  & 1.20       & 77.68 & 68.11 & 12.36 & {1.75}     & 591.87 & 61.99 \\
UNETR++ \cite{shaker2024unetr++} & 91.12 &  83.89 & 5.56 & 1.52 & 83.32 & 71.99 & 5.96 & \underline{1.05} & 79.45 & 70.00 & 11.84 & \underline{1.69} & 81.64 & 20.33 \\
MedNeXt  \cite{roy2023mednext}           & \underline{92.05} & \underline{85.35} & \underline{4.95} & \textbf{1.36}  & \underline{83.82} & \underline{72.50} & \underline{4.72} & 1.16  & \underline{82.88} & \underline{73.75} & 9.84 & 1.76        &  65.62 & 5.57  \\
\textbf{Light-UNETR}   & 91.58 & 84.51 & \underline{4.95} & \underline{1.43}  & 80.45 & 67.96 & 6.19  & 1.30           & 79.93 & 69.33 & 11.44 & 2.26    &  \textbf{4.29} & \textbf{1.34} \\
\textbf{Light-UNETR-L} & \textbf{92.22} & \textbf{85.61} & \textbf{4.64} & \textbf{1.36}  & \textbf{84.30} & \textbf{73.12} & \textbf{4.36} & \textbf{1.04}   & \textbf{84.32} & \textbf{74.98} & \textbf{9.15} & \textbf{1.65}  &  \underline{59.45} & \underline{3.93} \\
\bottomrule
\end{tabular}
} 
\end{table*}

Our proposed CSE-Light-UNETR method also achieves promising performance on the Pancreas-CT dataset in Tab. \ref{tab:merged-results}, outperforming state-of-the-art methods in both 6-labeled (10\%) and 12-labeled (20\%) scan scenarios. When using 6 labeled scans, our method surpasses the best-performed method BCP \cite{BCP} by a margin of 4.84\% in terms of Dice score, achieving a score of 73.77\%. This demonstrates the effectiveness of our method in leveraging limited labeled data to achieve accurate segmentation, which is crucial in real-world clinical scenarios where annotated data is scarce. With 12 labeled scans, our method outperforms MLRP \cite{su2024mutual_MLRP} by 6.70\% in Dice, achieving a Dice of 78.50\%. Notably, some methods that are designed for CNNs \cite{yu2019uncertaintyawaremeanteacher, wu2022ssnet} may perform worse than the model trained with only labeled data. This may be due to the fact that these methods leverage the inductive biases and local patterns that are inherent in convolutions and are not suitable for transformers that typically learn global representations. The observation further highlights the robustness of our approach. Overall, these results demonstrate the potential of CSE-Light-UNETR to improve the accuracy and reliability of pancreas segmentation in clinical practice. We also provide the qualitative segmentation results in Fig. \ref{fig:quali:pancreasct}. From the figure, we can see that most methods perform poorly under the hard pancreas segmentation task \cite{yu2019uncertaintyawaremeanteacher, wu2022ssnet, CAML}, while our proposed method could segment the anatomy accurately and pinpoint the error-prone regions. The results validate that the enriched contextual information by CSE could enhance the feature representation capability of the segmentation model.

\subsubsection{Results on BraTS 2019 dataset}

We evaluate our proposed method on the BraTS 2019 dataset for brain tumor segmentation, as shown in Table \ref{tab:results-brats}. 
Notably, our model with only supervised data could achieve 75.91\% Dice and 64.80\% Jaccard, demonstrating the strong baseline performance of our Light-UNETR model. With CSE, our model achieves a Dice score of 79.73\%, a Jaccard score of 68.76\%, a 95HD of 11.65, and with a significant reduction in ASD to 2.25. It outperforms CAML \cite{CAML}, BCP \cite{BCP}, and MLRP \cite{su2024mutual_MLRP} by a margin of 1.38\%, 2.67\%, and 1.63\% in terms of Dice score, respectively. Compared to the baseline with 75.91\% Dice and 3.86 ASD, it improves it by 3.82\% and 1.61. 
These results demonstrate the effectiveness of our method on varied scenarios.

\begin{table*}[h!]
\centering
\caption{\label{fullysupervised_part2}\xyl{Fully-supervised segmentation dice scores  on MSD Task01, AbdomenCT-1K, and HNC Tumor datasets. The best results are highlighted in \textbf{bold}, and the second-best are \underline{underlined}. FLOPs are computed with spatial size (96, 96, 96).}}
\resizebox{\textwidth}{!}{
\begin{tabular}{l|cccc|ccccc|ccc|cc}
\toprule
\multirow{3}{*}{{Method}} 
& \multicolumn{4}{c|}{{MSD Task01 dataset}} 
& \multicolumn{5}{c|}{{AbdomenCT-1K dataset}} 
& \multicolumn{3}{c|}{{HNC Tumor dataset}} 
& \multirow{3}{*}{{FLOPs (G)}} 
& \multirow{3}{*}{{Params (M)}}  \\
\cmidrule(lr){2-13}
& WT & ET & TC & Avg
& Liver & Kidney & Spleen & Pancreas & Avg 
& GTVp & GTVn & Avg 
& &
\\ 
\midrule
nn-UNet \cite{isensee2021nnunet}  & 85.83 & 62.71 & 83.42 & 77.32 & \underline{97.21} & \textbf{96.92}& \textbf{95.93}& 93.42& \textbf{95.87} & 94.79 & 95.76 & 95.27 & 200.81 & 13.82\\
SegResNet \cite{segresnet}  & 88.68 & 62.64 & 83.28 & 78.20 &97.06 &96.35 &95.23 &91.48&95.03 & 95.19 & 95.70 & 95.44 & 202.66 & 10.64 \\
TransBTS \cite{wang2021transbts}  & 77.89 & 57.44 & 73.51& 69.61 &97.08 &\underline{96.47} &95.38 &92.86&95.45& 94.02 & 94.31 & 94.16 & 142.07 & 30.61\\
Swin UNETR \cite{tang2022swinunetr}  & \underline{90.23} & 63.02 & 84.22&79.16 & 97.03 &96.08 &94.98 &\underline{93.51}&95.40 & 96.45 & \underline{96.51} & \underline{96.48} & 331.95 & 61.99\\
UNETR++ \cite{shaker2024unetr++} & 90.07 & 64.87 & 84.52& 79.82 & 95.64 & 95.56 & 93.99 & 91.68 & 94.22 & 95.39 & 95.74 & 95.56 & 57.51 & 20.33 \\
MedNeXt \cite{roy2023mednext} & 90.15 & {64.34} & {85.91} & {80.13}&  95.57& 95.30 &94.39 &91.32&94.15 & \textbf{96.62} & \textbf{97.03} & \textbf{96.82} & 57.81 & 5.57\\
\textbf{Light-UNETR} & \textbf{90.24} & \underline{67.01} & \underline{86.89}& \underline{81.38} & 94.36 &93.97 &92.90 &87.65 & 92.22 & 87.48 & 84.86 & 86.17 & \textbf{3.73} & \textbf{1.34} \\
\textbf{Light-UNETR-L} &{89.69}&\textbf{75.41}&\textbf{92.89}&\textbf{85.97}&\textbf{97.26} &96.38 &\underline{95.46} &\textbf{93.71}&\underline{95.70}  & \textbf{96.62} & 96.34 & \underline{96.48} & \underline{52.15} & \underline{3.93}\\
\bottomrule
\end{tabular}
} 
\end{table*}
\subsection{\xy{Fully-supervised Medical Image Segmentation}}
\subsubsection{\xy{Upper-bound Analysis for Semi-supervised Datasets}}
\xy{We first establish reference upper-bound results by evaluating Light-UNETR and Light-UNETR-L in the fully-supervised regime on the LA, Pancreas, and BraTS 2019 datasets (Table~\ref{fullysupervised_part1}). We compare with recent large-scale foundation models (MedSAM2~\cite{ma2025medsam2}, Efficient-MedSAM2~\cite{ma2025medsam2}, Med-FastSAM~\cite{luomedfastsam}) as well as popular 3D segmentation architectures (nnUNet~\cite{isensee2021nnunet}, SegResNet~\cite{segresnet}, TransBTS~\cite{wang2021transbts}, Swin UNETR~\cite{he2023swinunetr}, \xyl{UNETR++}~\cite{shaker2024unetr++} and MedNeXt~\cite{roy2023mednext}).
Light-UNETR-L consistently achieves the best Dice scores and boundary metrics on all three datasets. On BraTS 2019 dataset, it achieves 84.32\% Dice and 9.15 95HD with only 59.45G FLOPs, outperforms the second best methods MedNext by 1.44\% Dice and TransBTS by 0.46 95HD. Our lightweight Light-UNETR also delivers results competitive with much larger networks, while requiring significantly lower computational costs. Especially on the LA dataset, it achieves 91.58\% Dice and 1.43 ASD with merely 4.19G FLOPs and 1.34M parameters, which is competitive to segmentation models that are 30$\times$ larger. Compared to foundation models, they require substantially higher computational overhead due to their larger input resolutions for 3D inputs, yet deliver inferior segmentation accuracy compared to our method, which may be attributed to their lack of volumetric context exploitation and absence of medical-specific architectural biases. 
These results demonstrate the effectiveness of our architectural design, which effectively captures different frequency components within the input data, and fuse them in a computationally efficient manner. Our CGLU facilitates efficient channel communication and expressive nonlinear modeling with minimal parameter overhead.}

\begin{figure}[t]
    \centering
    \includegraphics[width=0.42\textwidth]{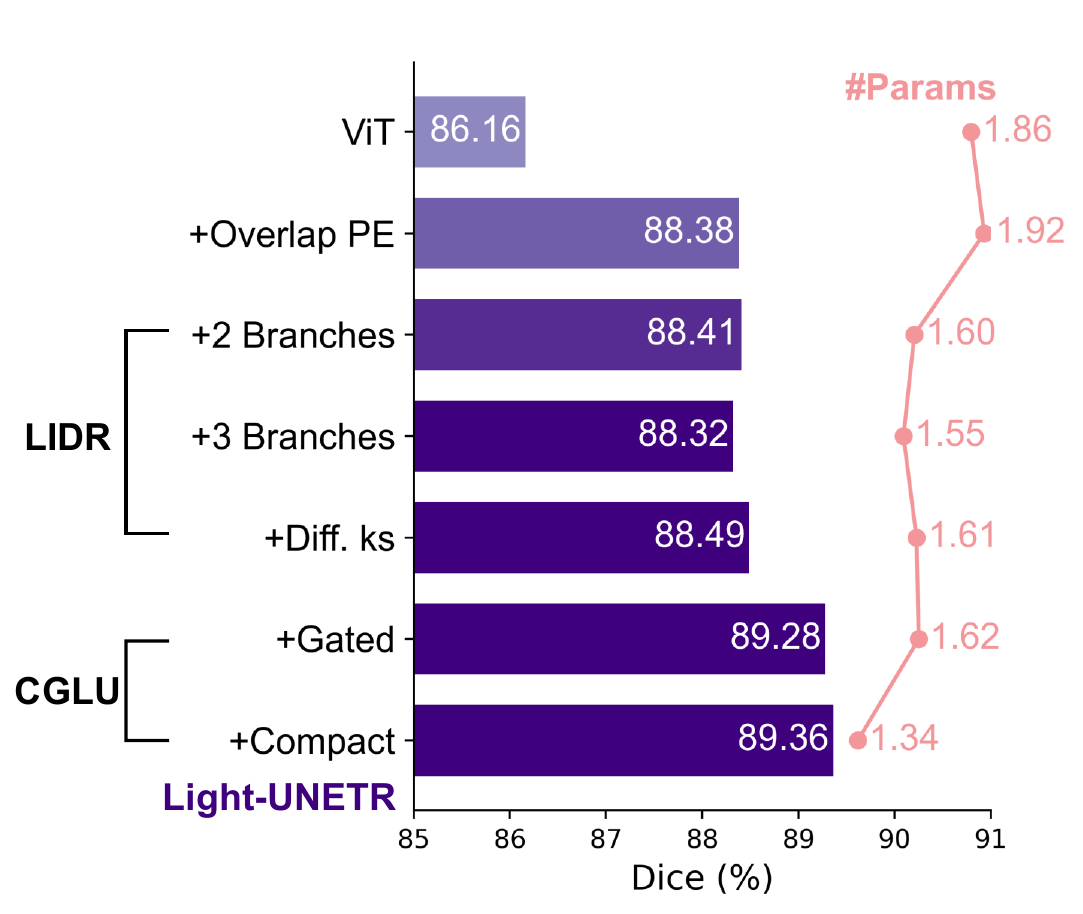}
    \caption{\xy{Evolution roadmap of the proposed Light-UNETR. The experiments are conducted on the LA dataset with 5\% labeled data.}}
    \label{fig:evo_road_ours}
\end{figure}
\subsubsection{\xy{Organ and Tumor Segmentation Performance}}
\xy{To comprehensively assess robustness and generalizability, we benchmark our models on three challenging fully-supervised segmentation tasks: MSD Task01 (brain tumor) \cite{decathlon}, AbdomenCT-1K (multi-organ abdomen) \cite{ma2021abdomenct}, and HNC Tumor (head \& neck cancer) \cite{HNCtumor}, detailed in Table~\ref{fullysupervised_part2}.
Light-UNETR-L again demonstrates excellent performance, achieving the highest or near-highest Dice scores on all datasets: an average Dice of {85.97\%} for MSD Task01, {95.70\%} on AbdomenCT-1K, and up to {96.48\%} on HNC Tumor. Compared to recent transformer and CNN-based baselines, Light-UNETR-L matches or exceeds accuracy while maintaining a dramatically smaller computational footprint and parameter set. The original Light-UNETR provides further evidence of the effectiveness and scalability of our design, delivering strong performance in a highly efficient regime with {3.73}~G FLOPs and {1.34}M parameters, which establishes a compelling trade-off for cost-sensitive deployment. \xyl{When compared to UNETR++ \cite{shaker2024unetr++}, our model outperforms it by 1.56\% in average Dice on MSD Task01, while significantly reduces the FLOPs by 93.5\%.}
In summary, both Light-UNETR and its large-scale variant Light-UNETR-L demonstrate that strong 3D segmentation performance does not require excessive model size or computational cost. Our design choices deliver the optimal efficiency-performance trade-off, robust generalization, and practical advantages for both research and clinical environments where hardware resources may be limited.}

\subsection{Ablation Study}

\begin{table}[t]
  \centering
  \caption{\label{comp_tabs}Ablation on the components in the proposed CSE. The experiments are conducted on the LA dataset with 5\% labeled data. Rand R.: Random Replacement. AGR: Attention-Guided Replacement. SMC: Spatial Masking Consistency.}
\setlength{\tabcolsep}{1.5mm}{\begin{tabular}[width=\textwidth]{cccc|cccc}
\toprule
\multicolumn{4}{c|}{Method} & \multicolumn{4}{c}{Metrics}  \\
\midrule
Sup Loss   & Rand R. & AGR   & SMC   & Dice$\uparrow$  & Jac$\uparrow$   & 95HD$\downarrow$  & ASD$\downarrow$  \\
\midrule
\checkmark   &       &       &       & 75.55 & 61.99 & 17.80  & 6.16 \\
\checkmark   &  \checkmark     &      &       & 86.78 & 77.35 & 8.37 & 2.84 \\
\checkmark   &       & \checkmark     &       & 87.79 & 78.97 & 7.51  & 2.33 \\
\checkmark   &       &       & \checkmark     & 83.59 & 72.60  & 11.45 & 3.79 \\
\checkmark   &       & \checkmark     & \checkmark     & 89.53 & 81.15 & 6.14  & 1.88 \\
\bottomrule
\end{tabular}}
\end{table}

\subsubsection{Evolution roadmap of Light-UNETR}
In Fig. \ref{fig:evo_road_ours}, we present the evolution roadmap of the proposed Light-UNETR. Notably, to avoid overfitting that may not truly reflect the performance, we split 10\% of the unlabeled training set of LA as validation set, and display the performance on the test set. Starting from a ViT model with stacked MHSA and FFNs \cite{ViT, chen2023vit3d, tang2022swinunetr}, with the overlap patch embedding (+Overlap PE), the model performance improved to 88.38\% with negligible extra parameters, demonstrating the effectiveness of low-level representations. Afterwards, we aim to reduce the computational cost of MHSA by gradually replacing it with LIDR. Specifically, when using low and high-frequency branches to learn both local and global information, the parameters are reduced by 0.26M, with a slight Dice improvement. With 3 branches, the number of parameters is further reduced to 1.55M, demonstrating the efficiency of using multiple frequency branches. With different kernel sizes for the two high-frequency branches, the performance and computational cost show a better trade-off. Next, we apply the components in CGLU to replace FFNs. With the gating mechanism, the performance is significantly improved to 89.28\% Dice with a slight 0.01M extra parameters, validating that effective channel communication is critical for lightweight vision transformers. Moreover, we adopt our compact design by reducing the expansion factor, the parameters are significantly reduced by 0.28M, and the Dice improved to 89.36\%. The results demonstrate that there exists parameter redundancy in existing Transformer design \cite{liu2023efficientvit}, and reducing it could lead to more efficient models without compromising performance, which may be attributed to the better propagation of information \cite{dai2018compressing} or avoiding overfitting to smaller-scale medical datasets \cite{lin2023lighter}. The results verify that Light-UNETR strikes a good balance between segmentation performance and model efficiency.

\begin{table}[t]
\setlength\tabcolsep{1.1pt} 
\centering
\caption{Comparison of $\lambda_{ext}$ and $\lambda_{int}$ settings (LA dataset, 5\% labeled data). Left: Equal $\lambda_{ext}=\lambda_{int}$; Right: Different $\lambda_{ext}\neq\lambda_{int}$.}
\renewcommand{\arraystretch}{1.2} 
\begin{tabular}{cccccc|cccccc} 
\toprule
$\lambda_{ext}$ & $\lambda_{int}$ & Dice↑ & Jac↑ & 95HD↓ & ASD↓ & $\lambda_{ext}$ & $\lambda_{int}$ & Dice↑ & Jac↑ & 95HD↓ & ASD↓ \\
\midrule
1.0 & 1.0 & 87.68 & 78.26 & 7.87 & 2.98 & 3.0 & 1.0 & 88.86 & 80.12 & 7.25 & 2.62 \\
2.0 & 2.0 & 88.45 & 79.45 & 6.86 & 2.39 & 4.0 & 1.0 & \textbf{89.53} & \textbf{81.15} & \textbf{6.14} & \textbf{1.88} \\
3.0 & 3.0 & 89.41 & 80.94 & 6.37 & 2.32 & 1.0 & 3.0 & 88.49 & 79.52 & 6.61 & 2.13 \\
4.0 & 4.0 & 89.41 & 80.93 & 6.28 & 2.31 & 1.0 & 4.0 & 88.31 & 79.22 & 9.60 & 3.02 \\
\bottomrule
\end{tabular}
\label{tab:lambda_hyperparams}
\end{table}

\begin{figure*}
    \centering
    \includegraphics[width=\textwidth]{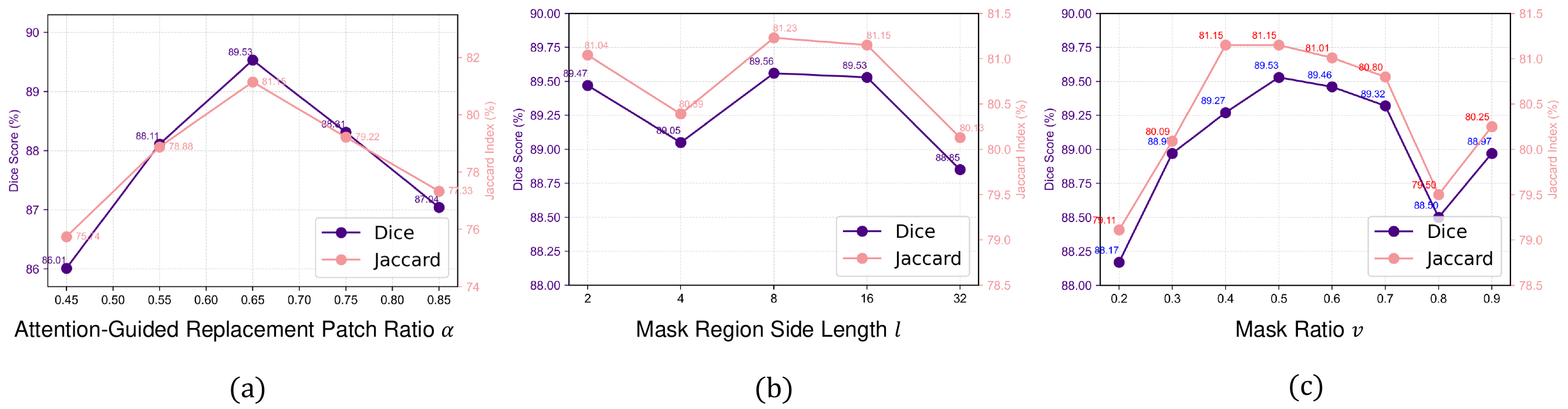}
    \caption{Ablation on the hyperparameters (a) The impact of changing the Attention-Guided Replacement patch size ratio $\alpha$. (b) The impact of changing the mask region side length $l$. (c) The impact of changing the mask ratio $v$.}
    \label{fig:enterhyperparamslabel}
\end{figure*}

\subsubsection{Ablation on the components in CSE}
To understand the improvements of the individual components in our proposed CSE, we conduct an ablation study on the LA dataset using 5\% labeled data and show the results in Tab. \ref{comp_tabs}. With only the labeled data and supervised loss, the model achieves a Dice score of 75.55\% and a Jaccard index of 61.99\%. Using random replacement can boost the Dice to 86.78\%, showing the importance of leveraging labeled data context to benefit unlabeled data learning. The incorporation of the Attention-Guided Replacement module leads to a significant performance boost, with the Dice score increasing to 87.79\% and the Jaccard index reaching 78.97\%. This highlights its effectiveness in enhancing the model's ability to learn from the limited labeled data via extrinsic context with appropriate guidance. Additionally, the inclusion of the Spatial Masking Consistency also improves the performance, with the Dice score and Jaccard index increasing to 83.59\% and 72.60\%, respectively. Finally, when all the proposed components including the Sup Loss, Attention-Guided Replacement, and Spatial Masking Consistency are integrated, the model achieves the best performance, with a Dice score of 89.53\% and a Jaccard index of 81.15\%. This demonstrates the synergistic effect of the individual components and their combined contribution to the overall model effectiveness.

\subsubsection{Ablation on the hyperparameters}
To better understand the impact of different hyperparameters on the model's performance, we conduct an ablation study and illustrate the results in Fig. \ref{fig:enterhyperparamslabel}. In Fig. \ref{fig:enterhyperparamslabel}(a), we test the different ratios $\{0.45, 0.55, 0.65, 0.75, 0.85\}$. The dice score initially improves, reaching a peak at 0.65, and then starts to decline. Similarly, the Jaccard index follows a similar trend, with the highest value at a replacement ratio of 0.65. The results suggest that selecting a proper ratio of extrinsic information for assisting the unlabeled data learning is critical for the performance. Then, we alter the mask region side length $l$ in Spatial Masking Consistency and display the results in Fig. \ref{fig:enterhyperparamslabel}(b). The model performance is relatively stable across different mask region side lengths, showcasing the robustness to masked sizes. Both side lengths 8 and 16 show slightly superior dice scores and Jaccard indexes, thus we utilize 16 in all our experiments. Lastly, we evaluate the effect of changing the mask ratio $v$ in SMC in Fig. \ref{fig:enterhyperparamslabel}(c), which controls the strength of intrinsic context learning. As the mask ratio increases from 0.2 to 0.9, the dice score and Jaccard index both improve at the beginning, reaching their peak values at a mask ratio of 0.5. Beyond this point, the performance starts to decline. Thus we set $v$=0.5 for the optimal performance.

\begin{table}[t]
\centering
\caption{\xy{Ablation on the number of high-frequency branches and kernel size configurations.}\label{tab:ablation_branch_ks}}
\resizebox{0.48\textwidth}{!}{
\begin{tabular}{cc|cc|cc|cc}
\toprule
\multirow{2}{*}{\#Branch} & \multirow{2}{*}{Kernel} & FLOPs & Params & \multicolumn{2}{c|}{LA (8/72)} & \multicolumn{2}{c}{Pancreas (12/50)} \\
\cmidrule{5-8}
 &  & (G) & (M) & Dice(\%) & Jac(\%) & Dice(\%) & Jac(\%) \\
\midrule
1 & 3 & 4.29 & 1.34 & 89.39 & 81.82 & 78.04 & 65.52 \\
1 & 5 & 4.30 & 1.43 & 89.33 & 81.83 & 77.69 & 65.30 \\
1 & 7 & 4.32 & 1.64 & 90.24 & 82.51 & 77.32 & 65.02 \\
2 & 3,5 & \textbf{4.29} & \textbf{1.34} & \textbf{90.50} & \textbf{82.74} & \textbf{78.50} & 65.27 \\
2 & 5,7 & 4.31 & 1.56 & 89.54 & 81.92 & 78.46 & \textbf{65.81} \\
3 & 3.5,7 & 4.30 & 1.46 & 90.19 & 82.24 & 78.38 & 65.23 \\
\bottomrule
\end{tabular}}
\end{table}
As shown in Table \ref{tab:lambda_hyperparams}, we also analyze the sensitivity in terms of the balancing terms for the learning of extrinsic context $\lambda_{ext}$ and intrinsic context $\lambda_{int}$. We first try a group of consistent parameters {1.0, 2.0, 3.0, 4.0} for both hyperparameters, finding that increasing the values leads to a performance improvement. By fixing $\lambda_{int}$, increasing $\lambda_{ext}$ boosts the overall performance, demonstrating that $\lambda_{ext}$ = 4.0 is optimal. By fixing $\lambda_{ext}$, adjusting $\lambda_{int}$ shows a slight impact on the framework. With the experiments, we find that setting $\lambda_{ext}$ = 4.0  and $\lambda_{int}$ = 1.0 brings the optimal performance. These results demonstrate that the larger extrinsic information contributes to the learning of semantics while extra intrinsic information exploration brings further improvements.

\subsubsection{\xy{Ablation on the number of branches and different convolution kernel-size combinations}}

\xy{we conduct ablation experiments in Tab. \ref{tab:ablation_branch_ks} to systematically evaluate the effect of varying the number of branches and different convolution kernel-size combinations. The experiments indicate that adopting two high-frequency branches with different kernel sizes (3 and 5) significantly improves segmentation accuracy compared to using single-branch designs with high efficiency, demonstrating that our design strikes an optimal balance between efficiency and representation capacity.}

\subsubsection{\xy{Ablation on the shared weight and EMA update}}

\xy{In SSL, both shared weight \cite{sohn2020fixmatch, liu2024most} and EMA updates \cite{meanteacher} are widely used strategies. In CSE, we propose the use of shared weights alone. To verify its effectiveness, we compare our model with EMA only and combining both shared weights and EMA, as shown in Tab. \ref{tab:ema}. The results demonstrate that shared weights not only reduce computational cost but also improve performance. The efficiency is due to its utilization of a single set of network weights, instead of requiring a separate teacher network. Moreover, we attribute the performance gains to the fact that medical image segmentation critically depends on precise boundary learning. Shared weights enable the model to receive immediate gradient feedback during each backpropagation step, allowing it to rapidly adjust and refine boundary details, particularly during the early stages of training. In contrast, EMA utilizes a historical average of model weights to smooth training, which can introduce a delay in adapting to new data and potentially obscure crucial boundary information, ultimately resulting in lower segmentation performance.}

\begin{table}[t]
    \centering
        \caption{\xy{Performance comparison between ours using shared-weight, using shared-weight and EMA simultaneously, and EMA only.}}
    \label{tab:ema}
\resizebox{0.48\textwidth}{!}{
    \begin{tabular}{lccccc}
        \toprule
        {Method} & Dice$\uparrow$ & Jac$\uparrow$ & 95HD$\downarrow$ & ASD$\downarrow$& Mem (GB) $\downarrow$\\
        \midrule
        Ours & \textbf{89.53} & \textbf{81.15} & \textbf{6.14} & \textbf{1.88} & \textbf{2.52}  \\
        Ours+EMA             & 86.85          & 76.96          & 8.73           & 2.32       & 3.10   \\
        EMA Only       & 84.53          & 73.70          & 10.21          & 2.82        &  3.10 \\
        \bottomrule
    \end{tabular}}
\end{table}

\begin{table}[t]
  \centering
  \caption{Segmentation performance comparison with CNN-based methods on the LA dataset with 10\% labeled data. FLOPs are computed with input size (112, 112, 80). $^{\dagger}$ refers to models using V-Net \cite{vnet} as base segmentation model.}
\setlength{\tabcolsep}{1.5mm}{\begin{tabular}[width=\textwidth]{c|cc|cccc}
\toprule
\multirow{2}{*}{Methods} & FLOPs & Params & \multicolumn{4}{c}{Metrics}                 \\
\cmidrule{4-7}
                 & (G) & (M)           & Dice$\uparrow$ & Jac$\uparrow$ & 95HD$\downarrow$ & ASD$\downarrow$ \\
\midrule
V-Net$^{\dagger}$ \cite{vnet}  & 46.85 &          9.44         & 82.74& 71.72& 13.35& 3.26  \\
UA-MT$^{\dagger}$ \cite{yu2019uncertaintyawaremeanteacher}     & 46.85 &          9.44                 & 87.79 & 78.39 & 8.68 & 2.12     \\
SS-Net$^{\dagger}$\cite{wu2022ssnet}      & 50.71 &          9.45             &  88.55 & 79.62 & 7.49 & 1.90        \\
CAML$^{\dagger}$   \cite{CAML}      & {119.72} & 19.73            & 89.62 & 81.28 & 8.76 & 2.02       \\
BCP$^{\dagger}$    \cite{BCP}      & 46.85 &          9.44                  &  {89.62}&{81.31}&{6.81}&\textbf{1.76}     \\
MLRP$^{\dagger}$ \cite{su2024mutual_MLRP} & 96.19 & 12.34 & 89.86 &81.68& 6.91&1.85\\
CSE-VNet$^{\dagger}$ & 46.85 &          9.44& 90.28&82.40&5.94&1.82 \\
CSE-Light-UNETR     & {4.29} & {1.34}       & \textbf{90.50}  & \textbf{82.74}  & \textbf{5.82}   & {1.83}      \\
\bottomrule
\end{tabular}}
\label{tab:compare-cnn}
\end{table}

\subsection{Further Analysis}

\subsubsection{Comparison with CNN on SSMIS}
To demonstrate the potential of our model for SSMIS, as shown in Table \ref{tab:compare-cnn}, we present a comparison of the segmentation performance with previous methods that use CNN \cite{vnet} as base model. Our method outperforms state-of-the-art CNN-based methods \cite{yu2019uncertaintyawaremeanteacher, wu2022ssnet, CAML, BCP, su2024mutual_MLRP} on the LA dataset, with significant improvements. For instance, our CSE-Light-UNETR achieves a Dice score of 90.50\%, outperforming the previous best method, BCP \cite{BCP}, by 0.88\%. This suggests that our proposed transformer-based model Light-UNETR can be a powerful tool for SSMIS, offering a promising alternative to traditional CNN-based approaches. Furthermore, we also extend CSE to CNN-based models to build a CSE-VNet variant, by replacing the attention guidance with random patch replacement, the performance is 90.28\% and 5.94 95HD, outperforming MLRP by 0.42\% Dice and 0.97 95HD, demonstrating its efficacy.

\begin{table}[t]
\centering
\caption{\label{tab:abdomenct1k_semi}\xy{Semi-supervised segmentation performance on the AbdomenCT-1K dataset with 10\% labeled data. Dice scores (\%) for each organ and average performances are reported.}}
\resizebox{0.98\columnwidth}{!}{
\begin{tabular}{c|c|cccc|c}
\toprule
{Methods}    & {L/U} & {Liver} & {Kidney} & {Spleen} & {Pancreas} & {Avg} \\
\midrule
Light-UNETR (Ours)                       & 80/0 & 83.36 & 84.02 & 83.71 & 68.72 & 79.95 \\
UA-MT (MICCAI'19)  \cite{yu2019uncertaintyawaremeanteacher}        & 80/720 & 87.19 & 85.76 & 84.53 & 72.87 & 82.59 \\
SS-Net (MICCAI'22) \cite{wu2022ssnet}        & 80/720 & 88.62 & 86.44 & 85.92 & 74.15 & 83.78 \\
CAML (MICCAI'23)  \cite{CAML}         & 80/720 & 90.91 & 88.13 & 87.60 & 78.63 & 86.32 \\
BCP  (CVPR'23)   \cite{BCP}        & 80/720 & 93.22 & 91.92 & 89.68 & 85.06 & 89.97 \\
MLRP  (MedIA'24) \cite{su2024mutual_MLRP}         & 80/720 & 91.63 & 89.72 & 89.09 & 85.31 & 88.93 \\
CSE-Light-UNETR (Ours)                       & 80/720 & \textbf{93.93} & \textbf{93.09} & \textbf{92.41} & \textbf{86.73} & \textbf{91.54} \\
\midrule
Fully Supervised                             & 800/0  & 94.36 &93.97 &92.90 &87.65 &92.22 \\
\bottomrule
\end{tabular}}
\end{table}

\subsubsection{\xy{Semi-supervised learning on larger scale datasets}}

\xy{We conducted additional experiments on the AbdomenCT-1K dataset, employing a larger labeled/unlabeled data ratio (e.g., 80 labeled and 720 unlabeled cases) and present results in Table \ref{tab:abdomenct1k_semi}. It is demonstrated that our proposed CSE-Light-UNETR still achieves state-of-the-art performance with 91.54\% average Dice in such setting, surpassing the second best BCP by 1.57\%. Therefore, our method is robust even when larger labeled datasets are available. Despite the fully supervised methods remaining superior, our method shows competitive performance using only 10\% labeled data, which is crucial in scenarios where labeled data is scarce.}

\subsubsection{\xy{Comparison with sparse attention mechanisms}}
\xy{We have included additional experiments incorporating sparse attention mechanisms, such as Linformer \cite{wang2020linformer}, Longformer \cite{beltagy2020longformer}, and Nyström \cite{xiong2021nystromformer}, and compared them with our LIDR module and the vanilla transformer \cite{vaswani2017attention} As shown in Table~\ref{compare_sparase}, while sparse attention methods like provide slight reductions in computational cost and comparable segmentation accuracy, our proposed LIDR module achieves the best trade-off, which is 0.44G FLOPs and 1.33M params lower than Linformer. These results further validate the effectiveness of our approach over previous sparse attention designs.}

\begin{table}[t]
\centering
\caption{\label{compare_sparase}\xy{Comparison of different attention mechanisms.}}
\resizebox{\linewidth}{!}{\begin{tabular}{l|cc|cc|cc}
\toprule
\multirow{2}{*}{{Model}} & {FLOPs} & {Params} 
& \multicolumn{2}{c|}{{LA}} 
& \multicolumn{2}{c}{{Pancreas}} \\
\cmidrule{4-7}
 & (G) & (M) & 4/76 & 8/72 & 6/56 & 12/50 \\
\midrule
Vanilla \cite{vaswani2017attention} & 4.83   & 2.68   & 89.05   & 90.04    & \underline{73.75}         & \textbf{78.57}          \\ 
Linformer \cite{wang2020linformer} & \underline{4.73}   & \underline{2.67}   & 89.01   & 89.79    & 73.26         & 78.17          \\ 
Longformer \cite{beltagy2020longformer} & 4.82   & 2.68   & 88.94   & 90.35    & 73.67         & 78.24          \\ 
Nyström \cite{xiong2021nystromformer} & 4.83   & 2.69   & \underline{89.52}   & \textbf{90.72}    & 73.36         & 78.15          \\ 
LIDR     & \textbf{4.29}   & \textbf{1.34}   & \textbf{89.53}   & \underline{90.50}    & \textbf{73.77}         & \underline{78.50}          \\ 
\bottomrule
\end{tabular}}
\end{table}

\subsubsection{\xy{Relationship between performance and the number of labeled and unlabeled samples}}

\xy{We analyzed the performance with varied labeled ratios across three datasets in Tab. \ref{tab:increaselabnum}. It is observed that on all 3 datasets, a labeled ratio of around 40\% is sufficient to surpass the fully supervised performance. Specifically, in the LA dataset, with 40\% labeled samples, the Dice score achieves 91.64\%, higher than fully supervised 91.58\%. Similarly on Pancreas-CT and BraTS 2019, the Dice scores with 40\% labeled ratio are 80.00\% and 80.58\%, which are higher than fully supervised 78.95\% and 79.93\%. Notably, even with 20\% labeled data, the performance is already approximately near optimal, which suggests that our proposed CSE effectively improves the model’s capacity of capturing data structure and enhance segmentation performance.}

\begin{table}[t]
\centering
\caption{\label{tab:increaselabnum}\xy{Performance metrics on 3 different datasets with different labeled and unlabeled sample ratios.}}
\resizebox{\linewidth}{!}{\begin{tabular}{l|c|ccccc}
\toprule
{Dataset} & {L/U} & {Dice$\uparrow$} & {Jac$\uparrow$} & {95HD$\downarrow$} & {ASD$\downarrow$} & {Ratio} \\ 
\midrule
{LA} & 4/76 & 89.53 & 81.15 & 6.14 & 1.88 & 5\% \\ 
                         & 8/72 & 90.50 & 82.74 & 5.82 & 1.83 & 10\% \\ 
                         & 16/64 & 91.06 & 83.40 & 5.16 & 1.70 & 20\% \\ 
                         & 32/48 & \textbf{91.64} & \textbf{84.61} & \textbf{4.81} & \underline{1.51} & 40\% \\ 
                         & Fully Supervised & \underline{91.58} & \underline{84.51} & \underline{4.95} & \textbf{1.43} & 100\% \\ 
\midrule
{Pancreas-CT} & 6/56 & 73.77 & 59.44 & 11.77 & 3.76 & 10\% \\ 
                  & 12/50 & 78.50 & 65.27 & 8.35 & 2.24 & 20\% \\ 
                  & 24/38 & \textbf{80.00} & \textbf{67.15} & \textbf{6.64} & \underline{1.48} & 40\% \\ 
                  & Fully Supervised & \underline{78.95} & \underline{66.17} & \underline{9.34} & \textbf{1.24} & 100\% \\ 
\midrule
{BraTS 2019} & 25/225 & 79.73 & 68.76 & 11.65 & 2.25 & 10\% \\ 
               & 50/200 & 79.74 & 68.92 & \underline{10.88} & \textbf{1.97} & 20\% \\ 
               & 100/150 & \textbf{80.58} & \textbf{69.84} & \textbf{10.61} & \underline{2.04} & 40\% \\ 
               & Fully Supervised & \underline{79.93} & \underline{69.33} & 11.44 & 2.26 & 100\% \\ 
\bottomrule
\end{tabular}}
\end{table}

\begin{figure}[t]
    \centering
    \includegraphics[width=0.46\textwidth]{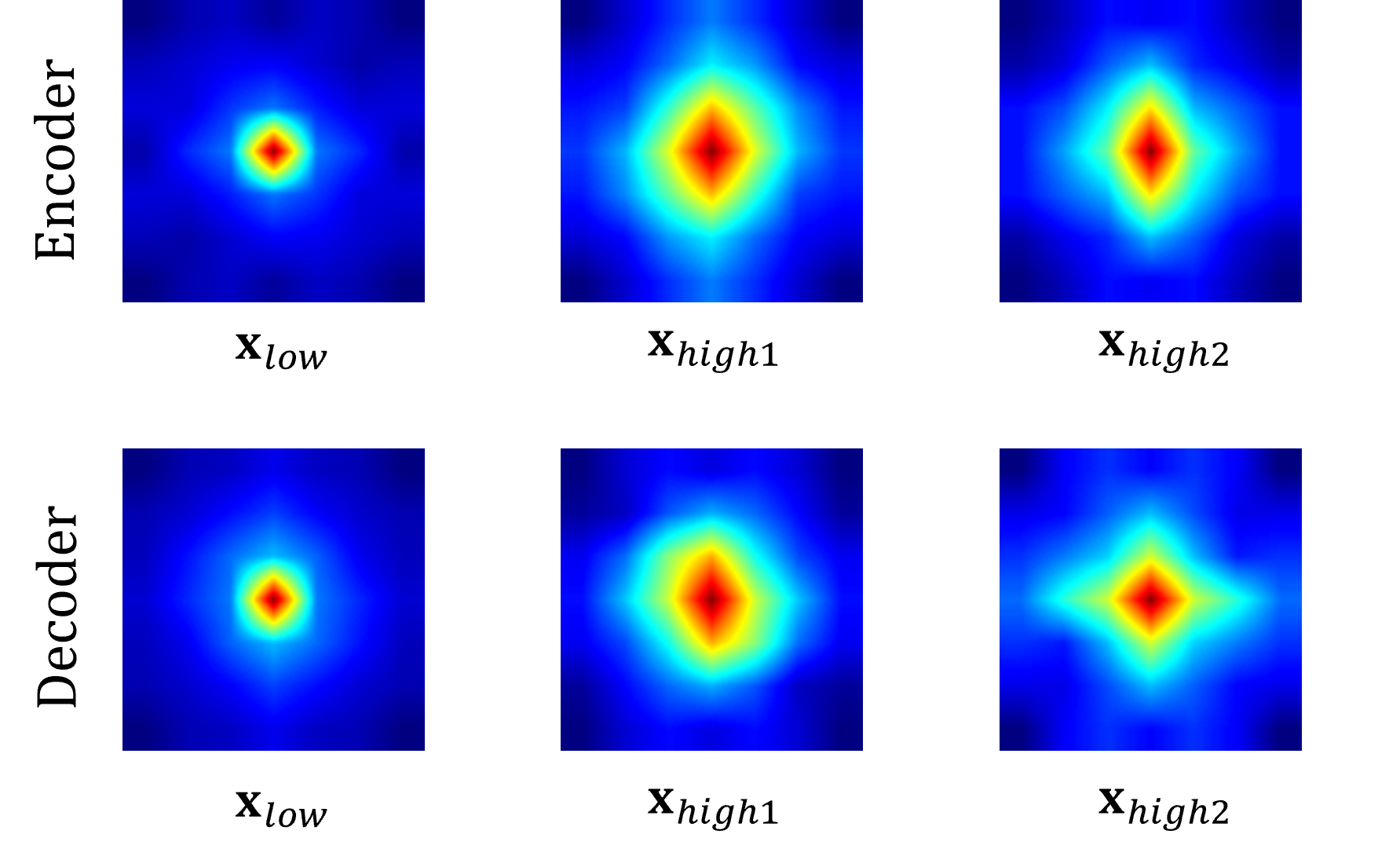}
    \caption{Fourier spectrum of the low-frequency branch and the high frequency branches. Top row: The Fourier spectrum of the feature in the encoder. Bottom row: The Fourier spectrum the feature in the decoder.}
    \label{fig:freq_vis}
\end{figure}

\subsubsection{Fourier Spectrum analysis for the LIDR module}

Our proposed LIDR module aims to extract multi-frequency components in the input medical data by low-frequency branch via self-attention and two high-frequency branches with convolutions with different kernel sizes. To further validate the effect of this design, we use Fourier spectrum analysis \cite{pan2022hilo, si2022inceptiontransformer} for the visualization of the encoder and decoder features, as displayed in Fig. \ref{fig:freq_vis}. It is observed that the branch with self-attention are more activated at the low frequency areas, while the convolution branches tend to activate more region with high frequency. Notably, the two branches with different kernel sizes capture distinct scale information within the medical volume, thereby improving the ability and robustness for feature extraction. Therefore, the results proves the perception capability of the proposed LIDR in the frequency spectrum.

\begin{table}[t]
\centering
\caption{\xyl{Compatibility of MedNeXt and Light-UNETR with hardware optimizations on fully-supervised LA dataset.}}
\label{tab:hardware}
\resizebox{0.48\textwidth}{!}{
\begin{tabular}{l|cccccc}
\toprule
{Model} & Dice (\%) & FPS $\uparrow$ & Mem (GB) $\downarrow$ & Params (M)\\
\midrule
MedNeXt \cite{roy2023mednext} & 92.05 & 28.43 (1.00x) & 12.50 (1.00x) & 5.57 \\
~ w/ FA            & 92.05 & 28.43 (1.00x) & 12.50 (1.00x) & 5.57 \\
~  w/ AMP           & 92.03 & 35.25 (1.24x) & 10.25 (0.82x) & 5.57 \\
~  w/ TensorRT      & 92.05 & 54.59 (1.92x) & 9.38 (0.75x) & 5.57 \\
\midrule
Light-UNETR            & 91.58 & 55.18 (1.00x) & 2.52 (1.00x) & 1.34 \\
~  w/ FA      & 91.58 & 67.07 (1.22x) & 2.48 (0.98x) & 1.34 \\
~  w/ AMP     & 91.57 & \underline{83.06} (1.51x) & \textbf{1.97} (0.78x) & 1.34 \\
~  w/ TensorRT& 91.58 & \textbf{104.29} (1.89x) & \underline{2.06} (0.82x) & 1.34 \\
\bottomrule
\end{tabular}
}
\end{table}
\begin{table}[t]
\setlength\tabcolsep{4.5pt}
  \centering
  \caption{Segmentation performance comparison with different self-supervised pretraining techniques. Experimental results on the Synapse multi-organ CT dataset, all models are pretrained and finetuned for 150 epochs.}
  \renewcommand{\arraystretch}{1.33}
  \begin{tabular}{c|cccc}
    \toprule
    Method & \multicolumn{1}{c}{Dice$\uparrow$} & \multicolumn{1}{c}{Jac$\uparrow$} & \multicolumn{1}{c}{95HD$\downarrow$} & \multicolumn{1}{c}{ASD$\downarrow$} \\
    \midrule
    \multicolumn{1}{c|}{TransUNet-ViT-B-16 (MIM)} & 61.03 & 47.62 & 73.60 & 21.82 \\
    {TransUNet-ViT-B-16 (SMC)} & \textbf{62.22} & \textbf{48.94} & \textbf{72.19} & \textbf{20.17} \\
    \bottomrule
  \end{tabular}
  \label{tab:pretrain}
\end{table}

\subsubsection{\xyl{Compatibility with hardware optimizations}}
\xyl{Our Light-UNETR model is not only intrinsically efficient by design but also fully compatible with mainstream hardware-level optimizations, including Flash Attention (FA)~\cite{dao2022flashattention}, automatic mixed precision (AMP), and TensorRT kernel fusion. Since these optimizations cannot alter the FLOPs or model parameters, we measure and compare the frames per second (FPS) and GPU memory consumption. As summarized in Table~\ref{tab:hardware}, Light-UNETR delivers substantially higher efficiency than MedNeXt while maintaining comparable segmentation accuracy. In the vanilla setting, Light-UNETR attains 55.18 FPS with 2.52 GB peak memory and 1.34M params, better than MedNeXt with 28.43 FPS, 12.50 GB, and 5.57M params. Incorporating FA yields further gains for Light-UNETR (67.07 FPS, 2.48 GB), whereas MedNeXt does not benefit from this as it is a fully convolutional model. AMP boosts both models, with Light-UNETR reaching 83.06 FPS and 1.97 GB versus 35.25 FPS and 10.25 GB for MedNeXt. When incorporated with TensorRT optimization, Light-UNETR achieves up to 104.29 FPS, significantly faster than MedNeXt with 54.59 FPS. These results demonstrate the flexibility of Light-UNETR, ensuring efficient deployment across diverse hardware-accelerated AI infrastructures.}

\subsubsection{Relation to masked image modeling} The proposed spatial masking and masked image modeling (MIM) \cite{xie2022simmim, chen2023vit3d} are related as they both involve introducing masked regions in the images. However, their objectives differ significantly. MIM aims to predict or recover the image content in the masked regions based on a reconstruction loss, while our method takes a different approach by encouraging the model to learn to infer neighboring regions using contextual information in a smooth manner. Moreover, naive MIM pretraining is not sufficient to capture complex context dependencies \cite{hoyer2023mic}, thus the consistency objective with Spatial Masking Consistency is effective and practical for SSMIS. 
We are intrigued whether SMC is capable for self-supervised pretraining. We conduct an exploratory experiment similar to \cite{chen2023vit3d} on TransUNet \cite{chen2021transunet} with ViT-B-16 backbone by pretraining and finetuning the model on Synapse dataset \cite{xu2016synapse}. Both the pretraining and finetuning phases consist of 150 epochs. We report the results in Table \ref{tab:pretrain}, despite SMC is not specially designed for pretraining, it achieves notably superior performance than the original masked image modeling and accelerates the fitting on downstream data \cite{tao2023siamese}. Concretely, it demonstrates substantial 1.19\%, 1.32\%, 1.41, 1.65 improvements in the four metrics, respectively. This suggests that SMC could be a viable alternative for pretraining, and opens up new possibilities for leveraging it in other medical imaging tasks.

\section{Conclusion}
\label{sec:conclusion}

The exploration of lightweight transformers in the domain of SSMIS has been a significant endeavor, given the computational efficiency, robustness, and superior performance required for clinical applications. This work first introduces Light-UNETR that aims to achieve model efficiency. Specifically, it is built upon a LIDR module, which is a self-attention mechanism that captures both global structural information and intricate local details through separate low and high-frequency branches, while requiring low computational costs by spatial and channel reduction. Meanwhile, a parameter-efficient module CGLU that selectively controls channel interactions is applied. We also scale it up to obtain a larger variant Light-UNETR-L. To deploy transformers to 3D medical image segmentation in a data-efficient manner, we opt to fully harness the unique characteristics of medical imaging data and explore the contextual relationships. Therefore, we propose a CSE framework, which is designed to integrate extrinsic contextual information via Attention-Guided Replacement and enhance intrinsic contextual reasoning via Spatial Masking Consistency. Experiments on diverse datasets demonstrate that our model and framework outperform state-of-the-art methods remarkably in both semi-supervised and fully-supervised settings in both efficiency and performance, highlighting the potential for real-world clinical applications where computational resources and labeled data are limited.
\bibliographystyle{IEEEtran}
\bibliography{mybib}

\vfill

\end{document}